\documentclass[lettersize,journal]{IEEEtran}
\usepackage{amsmath,amsfonts}
\usepackage{algorithmic}
\usepackage{array}
\usepackage[caption=false,font=normalsize,labelfont=sf,textfont=sf]{subfig}
\usepackage{textcomp}
\usepackage{stfloats}
\usepackage{url}
\usepackage{verbatim}
\usepackage{graphicx}
\usepackage{cite}
\usepackage{textcomp}
\usepackage[ruled,linesnumbered]{algorithm2e}
\usepackage{bm}
\usepackage{threeparttable}
\makeatletter
\newcommand{\removelatexerror}{\let\@latex@error\@gobble}
\makeatother
\usepackage{color}

\begin{document}
\title{A Feature Memory Rearrangement Network for Visual Inspection of Textured Surface Defects Toward Edge Intelligent Manufacturing}
\author{Haiming Yao, \IEEEmembership{Student Member, IEEE}, Wenyong Yu, \IEEEmembership{Member, IEEE}, and Xue Wang, \IEEEmembership{Senior Member, IEEE}
\thanks{Manuscript received XX XX, 20XX; revised XX XX, 20XX. This study was financially supported by the National Natural Science Foundation of China (Grant No. 51775214)
(\emph{Corresponding author: Wenyong Yu.})}
\thanks{Haiming Yao and Wenyong Yu are with the School of Mechanical Science and Engineering, Huazhong University of Science and Technology, Wuhan 430074, China (e-mail: ywy@hust.edu.cn; u201812016@hust.edu.cn).}
\thanks{Xue Wang is with the State Key Laboratory of Precision
Measurement Technology and Instruments, Department of Precision Instrument, Tsinghua University, Beijing 100084, China (e-mail:
wangxue@mail.tsinghua.edu.cn).}}

\maketitle

\begin{abstract}
  Recent advances in the industrial inspection of textured surfaces---in the form of visual inspection---have made such inspections possible for efficient, flexible manufacturing systems. However, establishing a unified manual-feature-based inspection model for homogeneous and nonregularly textured surfaces presents an enormous challenge. Furthermore, in real industrial scenarios, collecting and labeling sufficient defective samples is impracticable due to the scarcity of defects and the endless variety of defect types, thus limiting the performance of supervised deep learning methods. To address these challenges, we propose an unsupervised feature memory rearrangement network (FMR-Net) to accurately detect various textural defects simultaneously. Consistent with mainstream methods, we adopt the idea of background reconstruction; however, we innovatively utilize artificial synthetic defects to enable the model to recognize anomalies, while traditional wisdom relies only on defect-free samples. First, we employ an encoding module to obtain multiscale features of the textured surface. Subsequently, a contrastive-learning-based memory feature module (CMFM) is proposed to obtain discriminative representations and construct a normal feature memory bank in the latent space, which can be employed as a substitute for defects and fast anomaly scores at the patch level. Next, a novel global feature rearrangement module (GFRM) is proposed to further suppress the reconstruction of residual defects. Finally, a decoding module utilizes the restored features to reconstruct the normal texture background. In addition, to improve inspection performance, a two-phase training strategy is utilized for accurate defect restoration  refinement, and we exploit a multimodal inspection method to achieve noise-robust defect localization. We verify our method through extensive experiments and test its practical deployment in collaborative edge--cloud intelligent manufacturing scenarios by means of a multilevel detection method, demonstrating that FMR-Net exhibits state-of-the-art inspection accuracy and shows great potential for use in edge-computing-enabled smart industries.

\vspace*{10pt}

\emph{Note to Practitioners---}Most conventional visual inspection methods rely on supervised training and consequently require a large amount of labeled data and can detect only specific types of texture defects. In contrast, the proposed FMR-Net is a robust model for the simultaneous and accurate inspection of textured surfaces for various defects that does not require any real labeled defect samples. Furthermore, this model can also support a different fine-grained detection method that is very suitable in the edge computing paradigm. These two characteristics
are both extremely important for practical industrial applications. To the best of our knowledge, this is the first unsupervised edge intelligent vision inspection framework. As such, it can provide inspiration and serve as a reference for intelligent industry.
\end{abstract}


\section{Introduction}
\label{sec:introduction}

\IEEEPARstart{T}{he} detection of defects on textured surfaces is a crucial capability in industrial product quality control, with the aim of identifying regions with inferior profiles relative to the defect-free background. This capability is indispensable in many industrial scenarios involving various materials and products, such as fabric\cite{r1}, steel\cite{r2}, and thin film transistor (TFT)--light-emitting diode (LED) displays\cite{r3}. Therefore, to improve product quality and satisfy customers’ needs, surface defect detection has recently become increasingly important on industrial manufacturing lines.

\begin{figure}[t]
\centerline{\includegraphics[width=\columnwidth]{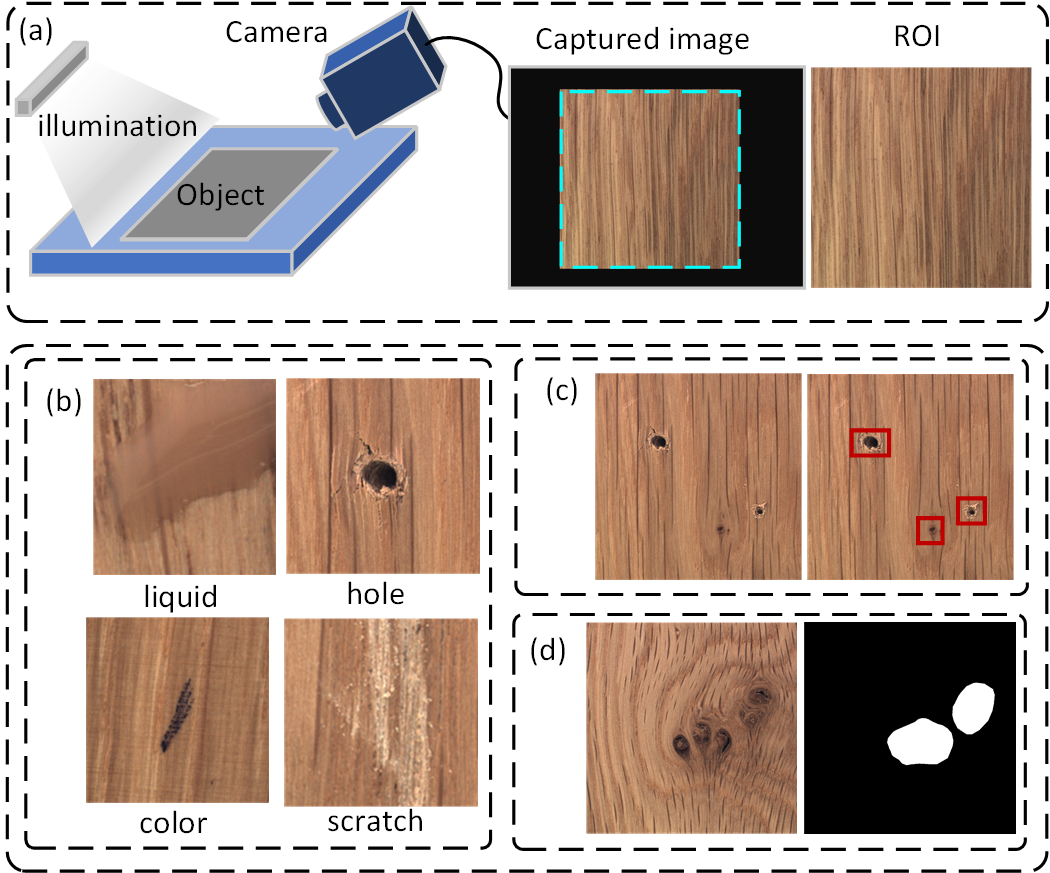}}
\caption{Overall schema of visual inspection for a textured surface. (a) Flowchart of image acquisition by AOI equipment. (b) Defect visual classification. (c) Defect object detection. (d) Defect segmentation.}
\label{fig1}
\end{figure}

In the context of industrial automation, visual inspection is a sensor-based, nondestructive technology that has been intensively researched due to its high accuracy and efficiency and has naturally become a widely deployed set of techniques in automated optical inspection (AOI)\cite{r4}. Generally, a visual inspection process follows the steps shown in Fig. \ref{fig1}. First, the textured surface to be inspected is imaged by an image sensor under preset illumination conditions. Then, the region of interest (ROI) in the captured image is obtained via conventional digital image processing techniques\cite{r5}. Subsequently, pattern recognition methods are employed to perform in-depth analyses, such as visual classification\cite{r6,r27}, object detection\cite{r7}, and segmentation\cite{r8,r54}. Among them, visual classification is an image-level process and thus cannot determine the defect location, which limits its practicality. While object detection technology usually produces a bounding box on each defect region as the identifier of that region, the precision of such technology still cannot meet the requirements of industrial scenarios. Therefore, the segmentation approach has become a popular focus of research, and many related achievements have been reported. At present, commonly used inspection methods can generally be divided into two categories: conventional methods and deep learning methods.

Conventional methods mainly utilize handcrafted features to obtain texture representations. In particular, researchers have investigated four subcategories of methods\cite{r9}: statistical schemes \cite{r10}, structural schemes\cite{r11}, spectral schemes\cite{r12}, and model-based schemes\cite{r13}. The phase-only transform (PHOT) was proposed by Aiger and Talbot\cite{r14} to detect defects as irregular mutations in background textures. Tsai and Huang\cite{r12} proposed low-pass filtering with curvature analysis (LCA) to remove periodic textural patterns. Xie\cite{r13} proposed texture exemplars (TEXEMS) for defect detection on randomly textured surfaces. These algorithms exhibit superior performance on specific types of materials, but until now, no unified handcrafted feature has been proposed that is applicable for multiple types of textured materials simultaneously.

Recently, deep learning has emerged as a trend in research aiming to improve the automatic extraction of discriminative and semantic features through the processing of big data. In the field of automatic visual inspection, deep learning approaches can be classified into supervised and unsupervised approaches based on their use of labeled defect samples. Supervised methods utilize a large number of defect samples and their corresponding labels to train a deep neural network (DNN). A two-stage network for the segmentation and classification of metal surface defects was proposed in \cite{r15}. \cite{r16} proposed pyramid feature fusion and a global context attention network to achieve defect segmentation. These methods provide a more flexible way to address inspection problems without using handcrafted features. However, a comprehensive defect data set is difficult to collect, and generating a large number of labels is usually impracticable in many production environments. As an alternative, unsupervised-learning-based methods require only defect-free samples and no labels for training. Mei \emph{et al}.\cite{r17} first proposed a multiscale convolutional denoising autoencoder (MSCDAE) to localize texture defects while using only defect-free samples for model training. Similarly, AnoGAN\cite{r18} utilizes the generator of a generative adversarial network to enable the detection of anomalies without the availability of defect samples in the training phase.

Currently, there is increasing interest in issues concerning the practical implementation of algorithms. Due to the heavy computational burden of DNNs, cloud computing, supported by strong computational facilities, has been widely used for calculation tasks in DNN applications\cite{r19}. Unfortunately, however, the nonnegligible latency due to data transmission has become a bottleneck for real-time inspection applications. To overcome this challenge, the newly developed edge computing paradigm \cite{r20} appears to offer a good solution for smart industrial scenarios.

We specify our basic assumptions as follows: Product defects are all caused by damage or abrasion either during the manufacturing process or during service. Thus, restoring a defective region to its proper textural profile while preserving the normal texture background can return a product to its initial configuration as a qualified product without defects. Given a pair of corresponding images, the defects can be located by comparing these two images.

The key to our approach is to identify and restore defects while maintaining the background texture profile. To this end, we propose a feature memory rearrangement network (FMR-Net) for the accurate inspection of textured materials for surface defects. Unlike in mainstream unsupervised methods, instead of using only defect-free images for training, we additionally employ artificial synthetic defects to simulate real defects. First, an encoding module extracts multiscale semantic information. Subsequently, a contrastive-learning-based feature memory (CMFM) is proposed for establishing a discriminative latent feature distribution and a normal texture feature memory bank, based on which defective latent features can be substituted to derive a restored representation and obtain a fast patch-level anomaly score. Next, the anomalous defective feature residuals engendered on skip pathways through multiscale feature extraction are suppressed by a newly proposed global feature rearrangement module (GFRM). Finally, a decoding module is employed to reconstruct the restored defect-less features into a normal texture background. Without any real defect samples, the proposed FMR-Net can suppress defect reconstruction to obtain an initial qualified texture configuration by means of a two-phase training strategy. Finally, in the testing phase, a novel multimodal inspection method based on isolated pixel residuals and local patch structures can reveal various defects simultaneously. More critically, based on an early exit mechanism\cite{r21} and model partitioning\cite{r22}, we propose and implement a multilevel detection method that can be applied in fast detection and cloud--edge collaboration scenarios, which shows great potential for use in smart industry.

The main contributions of our work are as follows:
\begin{enumerate}
\item We propose a novel feature memory rearrangement network (FMR-Net) from the perspective of defect restoration, a state-of-the-art model that can effectively perform an inspection of various types of textural defects simultaneously. The proposed methodology focuses on the following aspects:
\begin{itemize}
\item For the database, an artificial synthetic defect database is proposed for unsupervised-learning-based defect detection.
\item For the model, an anomaly feature editing technique named feature memory rearrangement is proposed for defect pattern recognition and restoration.
\item For the processing schema, we propose a novel two-phase training strategy and a multimodal inspection method for performance improvement.
\end{itemize}
\item From the perspective of the deployment schema, we implement the model in practice within the edge intelligent manufacturing paradigm. The proposed multilevel detection method utilizes an early exit mechanism and model partitioning to achieve fast detection and produce different fine-grained results, which can be exploited in cloud--edge collaboration mode.

\end{enumerate}

The remainder of this article is organized as follows. Section II provides additional background on the inspection of textured surfaces and edge intelligent manufacturing. FMR-Net is described in detail in Section III. Section IV presents extensive experiments conducted for performance verification and to test the edge intelligent implementation of FMR-Net. Finally, conclusions and discussions are given in Section V.

\section{Related Previous Works on the Defect Inspection of Textured Surfaces and Edge Intelligent Manufacturing}

\subsection{Defect inspection of textured surfaces}
Various approaches have been proposed for applying DNNs in unsupervised textured surface inspection, which is a topic of increasing interest in the computer vision field. To the best of our knowledge, the most prominent class of techniques for this purpose is deep generative models, which can be further categorized into two subtypes, autoencoders (AEs) and generative adversarial networks (GANs), along with their extensions.

An autoencoder (AE) is one type of unsupervised neural network in which the output target is to reconstruct the input data. In anomaly detection, the reconstruction error serves as a criterion for the anomaly property based on the fact that an AE accepts only normal samples as training data; thus, the reconstruction result for anomalous input should be unpredictable and yield a higher reconstruction error. However, this assumption may not always hold due to the strong generalizability of DNNs and the possibility that anomalies may be entangled with normal patterns in the latent space, in which case the AE will also prefer to reconstruct even anomalies well. To mitigate this drawback, many extensions have been proposed. The MSCDAE of Mei \emph{et al.}\cite{r17} utilizes a three-layer pyramid to perform textured surface defect inspection. Yang \emph{et al}.\cite{r9} proposed a multiscale feature-clustering-based fully convolutional AE (MS-FCAE), which focuses on constraining the feature distribution to eliminate defects during reconstruction. Gong \emph{et al}. \cite{r23} proposed MemAE, in which sparse coding of latent representations and memory-based reconstruction are introduced to eliminate abnormal generalization. \cite{r24} proposed a masked AE trained in a self-supervised fashion, which applies random removal and reconstruction of image regions through partial inpainting, thereby suppressing the generalization of the AE. However, a tradeoff exists between reconstruction quality for normal textures and defect suppression, limiting the defect inspection performance of AEs.

Recent developments in many visual tasks, such as texture synthesis\cite{r25} and image style transfer\cite{r26}, have shown that GANs have an outstanding generative ability and are highly capable of capturing the internal distributions of data. For reconstruction-based texture inspection, a GAN can leverage texture generation and restoration, in contrast to the simple texture reconstruction of AE-based models, to convert defective regions to the background texture profile.\cite{r28} proposed a GAN-integrated AE by replacing the decoder of an AE with the generator of a GAN. \cite{r29} proposed an anomaly feature-editing-based adversarial network (AFEAN) that suppresses the reconstruction of defects by means of latent feature editing. \cite{r30} proposed GANomaly, which jointly minimizes the distance between normal and pseudo samples in the image/latent domain; however, further experiments revealed that the clarity of the reconstructed image is limited in such an isolated bottleneck connected encoder--decoder network. To address this challenge, one natural approach named Skip-GANomaly was proposed in\cite{r31} by introducing `skip connections'. Although high-quality image reconstruction was achieved, unrepaired anomalous regions unfortunately remained due to the trivial solution of propagating low-level features directly to the last layers of the model. The AFEAN of\cite{r29} also ignored the impact of skip connections.

Overall, the phenomena of overinspection and misinspection remain as unresolved shortcomings in the above literature due to the following challenges. Only defect-free samples are employed for training, which limits the defect recognition ability of the model, and the tradeoff between anomaly elimination and high-quality reconstruction is difficult to balance. Moreover, inefficient unimodal defect localization techniques, such as naïve image subtraction, restrict the inspection performance.
\subsection{Edge intelligent manufacturing}
As a rapidly developing technology, edge computing\cite{r20} was proposed to enable the offloading of computing tasks to the edge of a communication network topology instead of transferring the data to the backend cloud. In the context of the Industrial Internet of Things (IIOT), edge intelligent manufacturing technologies have been developed for efficient computation and the continuous execution of manufacturing instructions. \cite{r32} proposed DeepIns for fog-computing-based industrial inspection. \cite{r33} proposed Boomerang for on-demand cooperative inference in the IIOT. However, only abstract concepts were presented in the above works; they included no practical deployment and considered only simple image classification tasks. A practical inspection system for turbo blades was proposed in\cite{r34} using the cloud--edge computing paradigm. A visual sorting system was proposed in \cite{r35} with the aid of edge computing, although these systems merely distributed the training and testing tasks. Therefore, the development and deployment of intelligent manufacturing systems within the IIOT paradigm remain challenging tasks.

In this paper, we propose FMR-Net to effectively inspect manufacturing products for various types of textural defects simultaneously by means of high-quality texture reconstruction and restoration. FMR-Net utilizes a CMFM for discriminative defect latent feature recognition and restoration and a GFRM to suppress residual defect reconstruction on skip connection pathways. Moreover, a two-phase training strategy and a multimodal inspection method are proposed for the reconstruction procedure and defect localization. Furthermore, we propose and
report the implementation of a multilevel inspection method that can provide different levels of effective fine-grained detection in cloud--edge collaboration mode. Subsequent parts of this article will provide more details.

\section{Proposed FMR-Net Methodology}

In this section, the proposed FMR-Net approach is introduced in detail. First, the overall framework of FMR-Net is briefly illustrated. Then, its main modules and contributions, including the artificial synthetic defect database, the contrastive-learning-based memory feature module (CMFM), and the global feature rearrangement module (GFRM), are presented in detail. Finally, the two-phase training strategy and multimodal inspection method are described.

\begin{figure*}[t]
\centerline{\includegraphics[width=\textwidth]{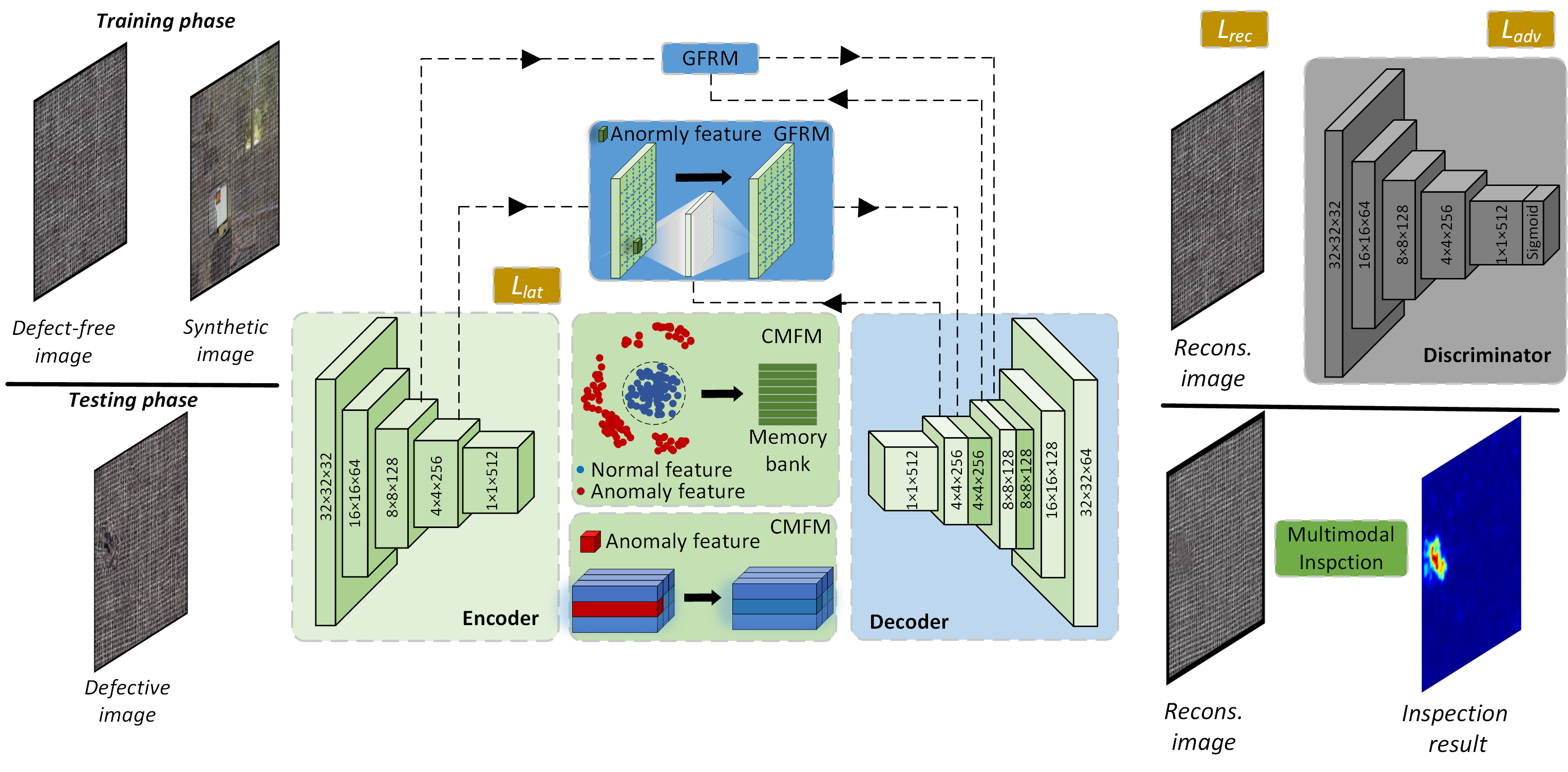}}
\caption[width=\textwidth]{
Overall architecture of the proposed FMR-Net method in the training and testing phases. FMR-Net adopts the GAN framework and consists of an encoder, a CMFM, a GFRM, a decoder, and a discriminator. During training, defect-free images and synthetic images are propagated forward, while real defective images are fed into the model only during testing. The brown boxes indicate losses. The encoder consists of five downsampling blocks, each composed of a convolutional layer with a 4×4 convolution kernel and a stride of 2, a batch normalization layer and a leaky ReLU activation layer. The upsampling blocks in the decoder have a similar composition. The latent features pass through the CMFM and then enter the decoder. The flows from the two deepest intermediate downsampling layers in the encoder and the corresponding upsampling layers in the decoder pass into the GFRM, where the outputs are concatenated to the corresponding layers in the decoder. The discriminator is identical to the encoder.

}
\label{fig2}
\end{figure*}

\subsection{Overall architecture of the FMR-Net model}

There are two key restrictions of reconstruction-based textural defect inspection methods. The first is that the current mainstream unsupervised deep learning models rely on only defect-free samples for training, resulting in a lack of defect recognition ability in the resulting model; thus, defects tend to remain unrepaired during reconstruction, leading to misinspection. The second is the stability of the normal background textural profile during the reconstruction procedure; if the normal texture is reconstructed imperfectly, pseudodefects or overinspection may occur. In this work, the novel FMR-Net model is proposed to address these shortcomings.

The overall architecture of FMR-Net is shown in Fig. \ref{fig2}. A GAN-integrated reconstruction model is employed as the basic pipeline. The generator network is composed of an encoding module, a CMFM, a GFRM and a decoding module. First, the encoding module extracts multiscale representative features of the input image. Then, the novel CMFM is applied for latent feature restoration. Based on contrastive learning, feature constraints in the CMFM can improve the discriminative ability of the features. The memory bank in the CMFM, which contains the typical normal texture mode, can be used to substitute for anomalous features and obtain anomaly scores at the patch level. Regarding the residuals of low-level intermediate defect features on the skip pathways, the novel GFRM is proposed to suppress them. Finally, the decoding module utilizes the restored defect-free features to reconstruct the input defect image into the background texture.

In the training stage, the whole model is optimized under the supervision of three losses, i.e., the latent loss ${L_{lat}}$, the reconstruction loss ${L_{rec}}$, and the adversarial loss ${L_{adv}}$. Two types of samples are used for training: defect-free images and artificial synthetic defect images used to simulate the defect mode. A novel two-phase training strategy, which includes multiple learning tasks, is proposed for hierarchical optimization of the model. In the first training phase, the model receives defect-free and synthetic defect samples as input. For the textural background reconstruction task, the encoding module, the decoding module and the GFRM are optimized under the supervision of the reconstruction loss ${L_{rec}}$ and the adversarial loss ${L_{adv}}$. For the latent feature distribution constraint task and the memory bank establishment task, the latent loss ${L_{lat}}$ is utilized to constrain the distribution of the latent features to have small intraclass distances and large interclass distances, thereby optimizing the parameters of the encoding module and the CMFM. Moreover, the feature memory bank is established during this phase. In the second training phase, under the supervision of the reconstruction loss ${L_{rec}}$ and the adversarial loss ${L_{adv}}$, the training of the decoding module,the CMFM and the GFRM is boosted by using artificial synthetic defect images to achieve the defect restoration task.

In the testing stage, the well-optimized FMR-Net is deployed for the defect inspection of textured surfaces. When a candidate defective image is fed into FMR-Net, the model can eliminate the anomalous defective regions, which are transformed into the background textural profile. For comparison-based defect localization, we propose a novel multimodal inspection method. Three different modalities are exploited to sufficiently explore the image characteristics on the basis of the independent pixel residuals and local structural differences. Thus, FMR-Net can be used to robustly inspect surfaces for textural defects.

\begin{figure}[t]
\centerline{\includegraphics[width=\columnwidth]{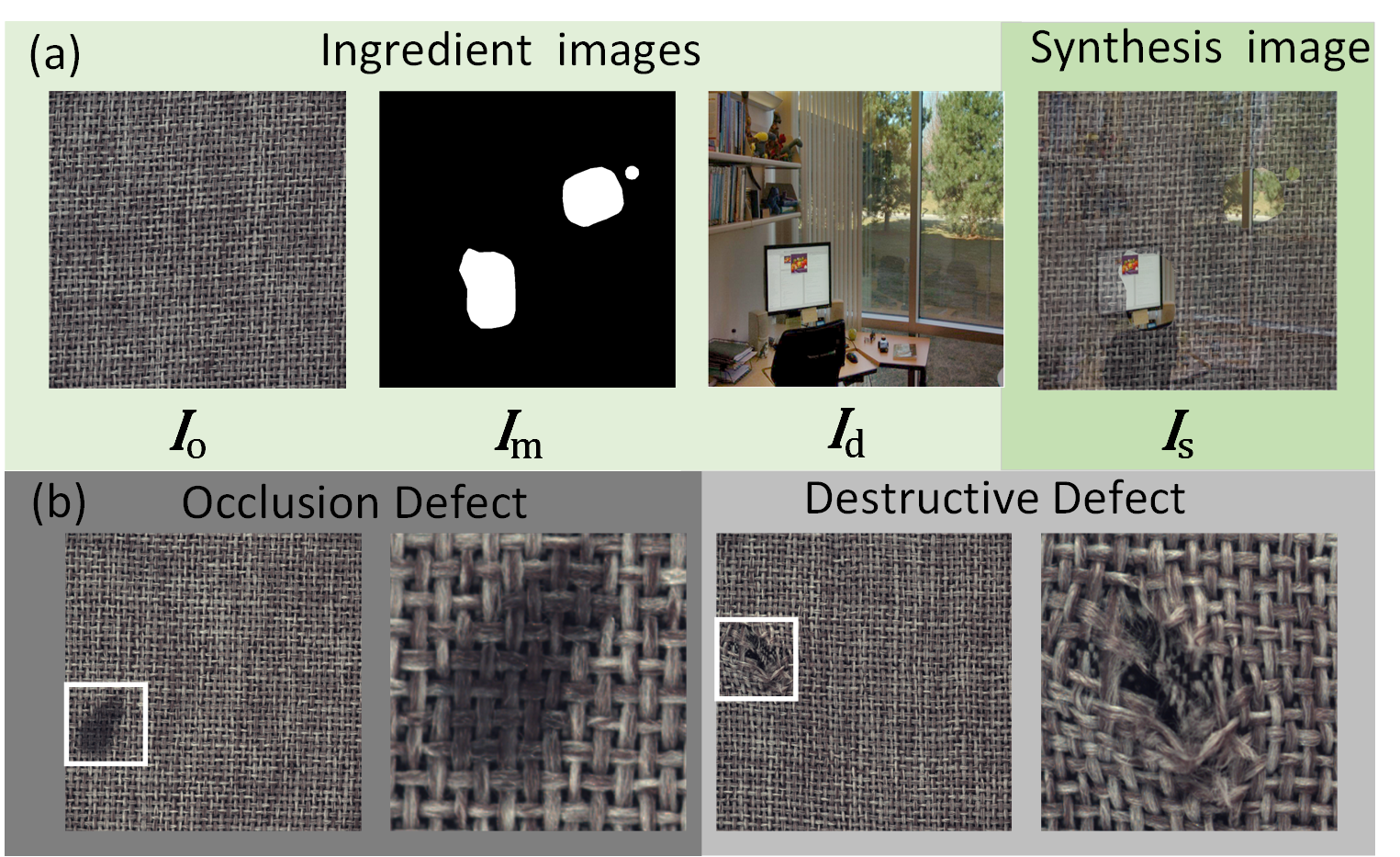}}
\caption{Interpretation of the image data set. (a) Formation process for artificial synthetic defect images. (b) Summary of typical real defect types.}
\label{fig3}
\end{figure}

\subsection{Artificial synthetic defect database}

As discussed above, background-reconstruction-based approaches can be summarized as consisting of two subtasks, i.e., the steady reconstruction of the normal background texture and the restoration of defect-degraded texture regions. The former problem has already been solved satisfactorily by AE-based models based on nondefect samples; however, attempts to solve the latter tend to be inefficient and coarse because anomalous regions are often also reconstructed with high fidelity.
In this paper, a novel artificial synthetic defect database is proposed, as shown in Fig. \ref{fig3} (a). We employ a defect-free image ${I_{o}}$, a random binary mask ${I_{m}}$, and a random anomalous image ${I_{d}}$ to synthesize the artificial defect image ${I_{s}}$:
\begin{equation}
{I_s} = \lambda  \times [(1 - {I_m}) \odot {I_o} + {I_m} \odot {I_d}] + (1 - \lambda ) \times {I_d}
\label{eq1}
\end{equation}
where ${\lambda  \in (0,1) }$ is the transparency, ${ \odot }$ is the elementwise product operation, and ${{I_s},{I_m},{I_o},{I_d} \in {R^{H \times W \times C}}}$, with $W$, $H$ and $C$ denoting the width, height, and number of channels, respectively, of the images. Moreover, we define two main categories of typical defects, namely, occlusion defects and destructive defects, as shown in Fig. \ref{fig3} (b). The former refers to regions overlaid with extraneous translucent substances, such as oil stains or water stains, while the latter refers to regions where the surface texture is entirely destroyed. Notably, the random anomalous images ${I_d}$ are sampled from natural images with abundant textural profiles, which can simulate various types of defects. The random mask ${I_m}$ is used to determine the defect category, i.e., an occlusion defect or a destructive defect. Through such defect simulation, the model is endowed with a defect recognition ability.

\subsection{Contrastive-learning-based memory feature module (CMFM)}

For a texture-reconstruction-based AE model, the encoder is used to obtain compressed representations that reveal the typical patterns of the input. As mentioned before, when the patterns of anomalies are entangled with those of normal textures, such a model also prefers to reconstruct anomalous defects with high fidelity. For a texture-generation-based GAN model, the textures are decoded from vectors with a certain distribution.
In the existing method of\cite{r30}, only normal textural images are mapped to this distribution, while the mapping regulation for anomalies is unpredictable, resulting in their latent features falling into regions that cannot be sampled. Hence, the implicit mapping process of the encoder should be elaborately predesigned. To overcome the related challenges, this section introduces the proposed CMFM for \emph{discriminative-feature-constrained encoder optimization} and \emph{feature-memory-based reasoning}.

\subsubsection{Discriminative-feature-constrained encoder optimization} For the regular training process of the encoder ${E}$, only a set of patches from normal images are utilized; consequently, the mapping results for defective samples are unpredictable. In contrast, we adopt triplets of image patches, each consisting of a sample patch ${P^0}$, a positive patch ${P^ + }$ and a negative patch ${P^ -}$, where ${P^0}$ and ${P^ + }$ are sampled from the defect-free image ${I_{o}}$ and ${P^ - }$ is sourced from the artificial defect image ${I_{s}}$. When anomalous images are introduced during training, the generation of discriminative latent representations is critical for anomalous defect identification. To this end, an auxiliary classifier ${C_{A}}$ (consisting of three fully connected layers with 512, 256, and 1 neurons and a sigmoid activation layer) is introduced for feature discrimination, and a novel latent loss function is proposed:

\begin{equation}
\begin{aligned}
L_{lat}^{hid}({R^0},{R^+},{R^-}) = {L_{BCE}}([{C_A}({R^ + }),&{C_A}({R^ - })],[1,0])\\
 &+ \left\| {{R^ + } - {R^0}} \right\|_{2}
\end{aligned}
\end{equation}
where ${R^0}$, ${R^+}$, and ${R^-}$ denote the latent feature vectors of ${P^0}$, ${P^ + }$, and ${P^ -}$, respectively, extracted by the encoder ${E}$ in the latent space. The last term is designed to minimize the Euclidean distance between ${R^ 0 }$ and ${R^ + }$ in the latent space, which encourages sparse regularization for generalization suppression\cite{r23}. Moreover, it forces the normal texture features to lie within a small data-enclosing hypersphere, which has been proven effective for texture defect detection in Deep Support Vector Data Description (Deep SVDD)\cite{r36}. The first term helps the auxiliary classifier ${C_{A}}$ distinguish abnormal features by performing feature classification based on the binary cross entropy (BCE):
\begin{equation}
{L_{BCE}}({\bf{x}},{\bf{y}}) =  - \left[ {{\bf{y}} \times \log ({\bf{x}}) + {\rm{1 - }}{\bf{y}}) \times \log (1 - {\bf{x}})} \right]
\label{eq1}
\end{equation}

\begin{figure}[t]
\centerline{\includegraphics[width=\columnwidth]{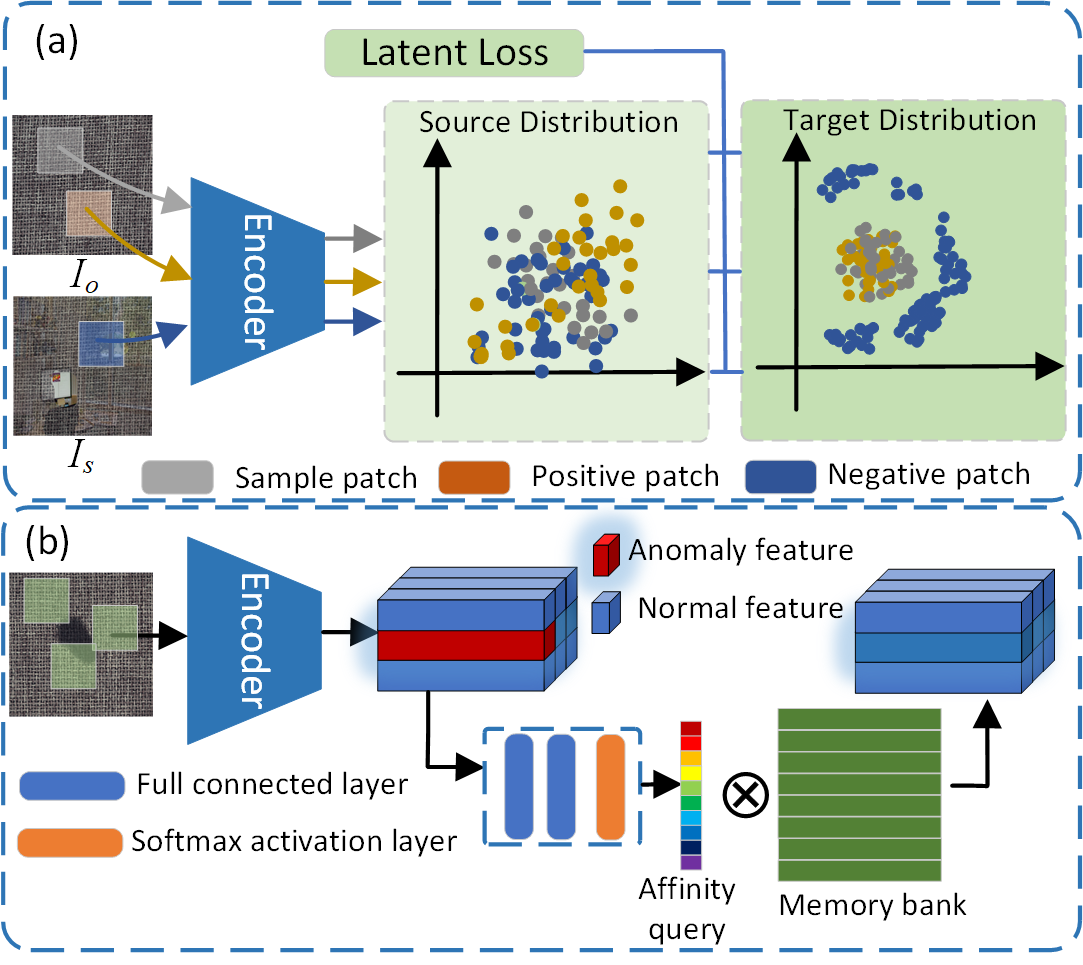}}
\caption{Schematic of the CMFM. (a) Discriminative-feature-constrained encoder optimization. The parameters of the CMFM and the encoder are updated based on ${L_{lat}}$ to make the original entangled distribution of the latent features approach the target discriminative distribution. The features are dimensionally reduced by means of t-distributed stochastic neighbor embedding (t-SNE). (b) Feature-memory-based reasoning. The normal latent features output by the encoder optimized in (a) are stored to form a memory bank. Then, anomalous latent features are propagated into the addressing network to form affinity queries to match reasonable entries in the memory bank, which are used to repair the corresponding defects.}
\label{fig4}
\end{figure}

The result for ${C_{A}}$ ranges from 0 to 1, and the labels of ${R^ + }$ and ${R^ - }$ are 1 and 0, respectively. As shown in Fig. \ref{fig4} (a), the source distribution corresponds to the unconstrained latent feature distribution, in which different types of features are entangled with each other, leading to high-fidelity defect reconstruction, whereas the target distribution is the latent distribution optimized by means of the latent loss function, which improves the discriminability of features and the sparsity of the normal features, both of which are beneficial for anomalous defect elimination.

\begin{figure}[t]
\centerline{\includegraphics[width=\columnwidth]{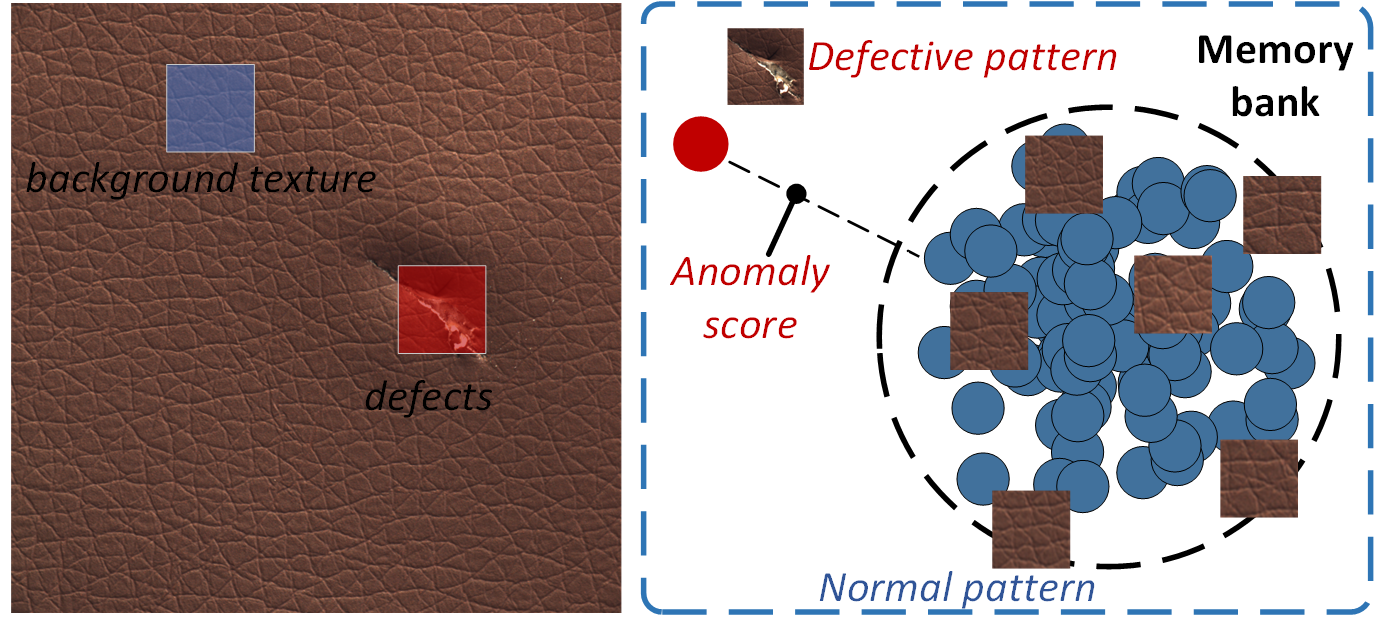}}
\caption{Schematic of the memory bank. The normal patterns corresponding to the background textures confined within the hypersphere are stored. The latent distance between a defective pattern and the nearest entry in the memory is defined as the patch-level anomaly score.}
\label{fig5}
\end{figure}

\subsubsection{Feature-memory-based reasoning} As shown in Fig. \ref{fig4} (b), after training, inspired by\cite{r37}, we utilize all encoded normal samples ${R^ +}=E({P^ +})$ in the training set to configure a memory bank $\bf{M}$ to record typical normal patterns. In the testing phase, a set of encoded abnormal features $z$ will be substituted by a linear combination of entries in $\bf{M}$. Unlike the similar method in\cite{r23}, our method uses a more flexible addressing scheme in a data-driven fashion. First, the encoded anomaly feature vector $Z$ is used to compute an affinity query $Q$ to match the most reasonable entries in $\bf{M}$ by an addressing network ${A_N}$ (consisting of two fully connected layers and a softmax activation layer with 256,512 neurons):
\begin{equation}
Q = {A_N}(z)
\label{eq1}
\end{equation}

Then, a linear combination of the most representative entries can be retrieved to the decoder:
\begin{equation}
\hat z = Q{\bf{M}} = {\sum\nolimits_{i = 1}^L Q _i}{M_i}
\label{eq1}
\end{equation}
where $Q \in {R^{1 \times L}}$, ${\bf{M}} \in {R^{L \times K}}$, $L$ denotes the number of memory entries in $\bf{M}$, $K$ represents the feature dimensions, $Q$ is a nonnegative row vector whose entries sum to 1,
${Q_{i}}$ represents the $i$th entry of $Q$, and $\hat z$ is the restored feature vector. Note that the hyperparameter $L$ is equal to the number of clusters\cite{r37}; we set its value to 512. Thanks to the enclosed hypersphere constraint based on $L_{lat}$, all possible combinations of memory entries will always lie inside the enclosed hypersphere, guaranteeing stable normality. Meanwhile, the neural network (NN)-based model ${A_N}$ provides a flexible, efficient addressing scheme.

Additionally, the latent distance between the encoded abnormal feature vector $z$ and the corresponding nearest neighbor entry in $M$ is regarded as the anomaly score for the corresponding patch, thereby enabling fast patch-level fine-grained anomaly judgment as shown in Fig. \ref{fig5}, which will be discussed in the specific case study presented later in this paper.

\subsection{Global feature rearrangement module (GFRM)}

\begin{figure}[t]
\centerline{\includegraphics[width=\columnwidth]{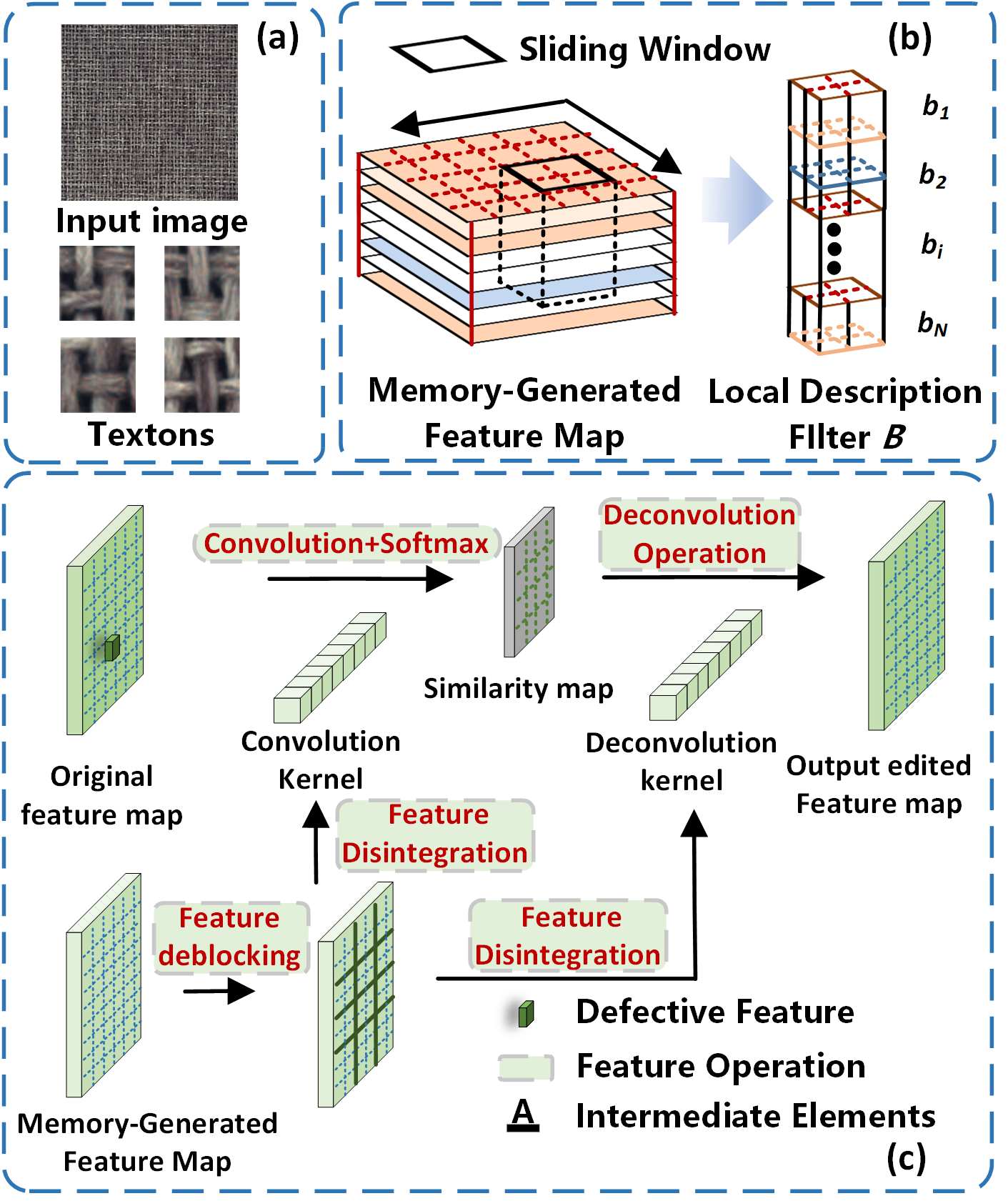}}
\caption{Architecture of the GFRM. (a) Textural image and its textons. (b) Flowchart of the formation process for deep feature textons (mainly divided into feature deblocking and feature disintegration). (c) Illustration of the GFRM.
}
\label{fig6}
\end{figure}

To pursue multiscale semantic information, skip connections are used to propagate features between the encoder and decoder at layers corresponding to various scales, which is beneficial for the accurate reconstruction of details. However, as discussed before, residual defect features may remain unrepaired as a result of this trivial solution. To address this issue, inspired by the concept of contextual attention in plausible image inpainting\cite{r38}, the novel GFRM is proposed.

\begin{figure}[t]
\centerline{\includegraphics[width=\columnwidth]{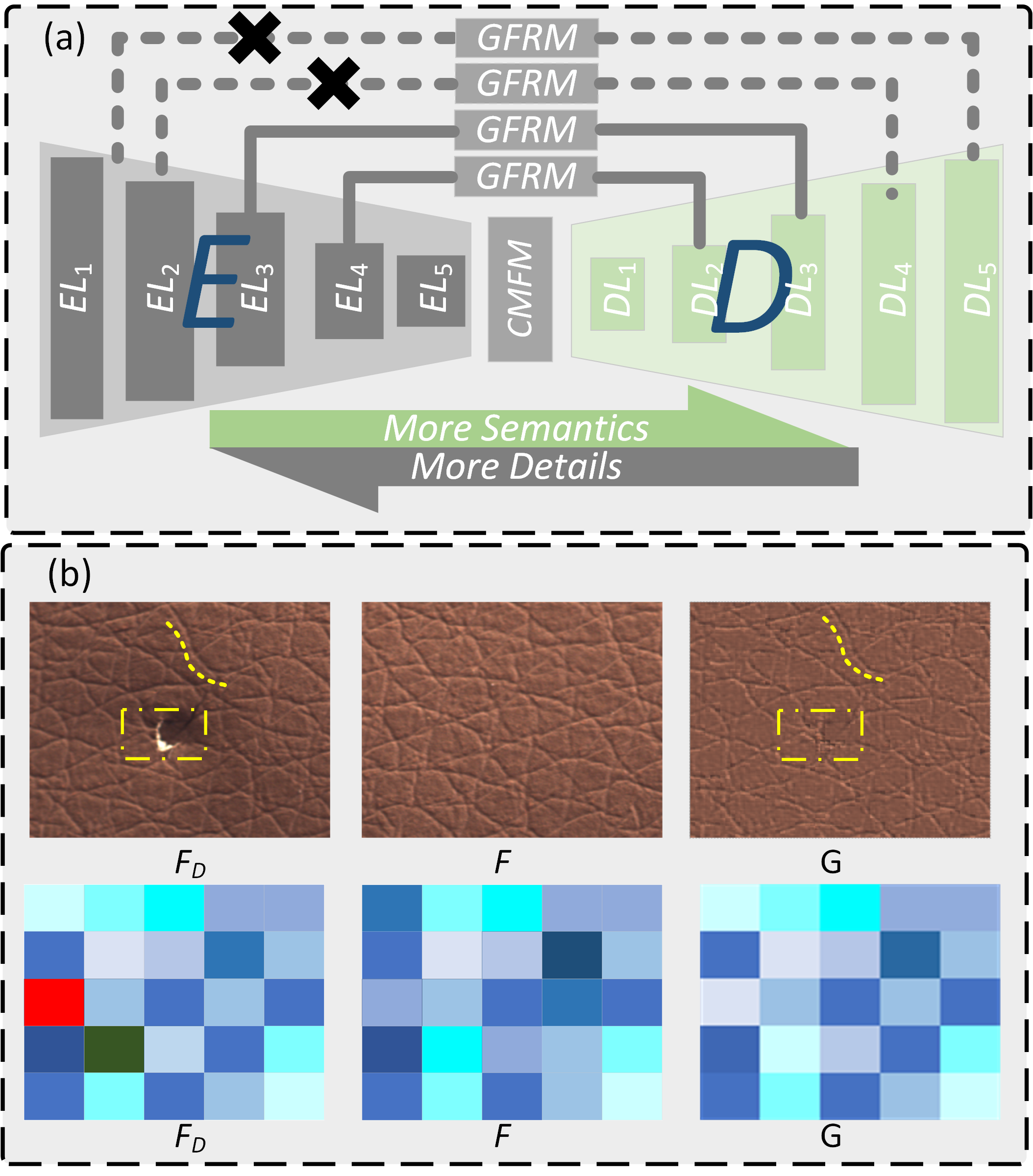}}
\caption{(a) Schematic of the GFRM embedding location.The EL and DL indicate the encoder layer and decoder layer respectively. (b)Effect of the GFRM. The first row shows the simulation of low-level features, the second row denotes the simulation of high-level features.}
\label{fig7}
\end{figure}

For residual anomaly feature suppression, we provide crucial insight into the rearrangement of the basic patterns of anomalies to match the typical microstructures of normal textures. In the field of texture analysis, the basic microstructures of typical textures are named textons\cite{r39}, which can be regarded as the basic typical local descriptors of a texture, as shown in Fig. \ref{fig6} (a). Due to the properties of convolutional neural networks (CNNs), the code vectors of the feature maps are semantically aggregated from the local regions of the input image, which contain all of the typical textural information; thus, they can naturally be viewed as textons. As shown in Fig. \ref{fig6} (b), the feature maps hierarchically upsampled from the CMFM, which are named memory-generated feature maps and are denoted by $F \in {R^{{H_F} \times {W_F} \times C}}$, can be decomposed into a set of textons $B = \{ {b_1},{b_2},...,{b_N}\} $, where ${b_i} \in {R^{K \times K \times C}}$, $i \in (1,...,\frac{{{H_F}}}{K} \times \frac{{{W_F}}}{K})$, with ${H_{F}}$ and ${W_{F}}$ being the feature map dimensions and $K$ being the texton size, which is set to 2. In accordance with the feature-memory-based reasoning procedure, the memory-generated feature maps are constrained within the hypersphere that contains the semantics of normal textures, thereby guaranteeing the normality of the textons. All of these textons are employed to form a local description filter. Notably, the parameters in a common CNN are trainable, whereas the parameters in the local description filter of the GFRM reveal the primitives of the texture, which are fixed. Intuitively, we can remove residual defects by adopting a linear combination of various textons for substitution, as shown in \ref{fig6} (c). Mathematically, we denote the original feature map on the skip pathway by ${F_D}$. To match defective patches with textons, we use the normalized similarity:
\begin{equation}
{{\bf{s}}_i}(p,q) = \frac{{\exp (d({F_D},{{{b}}_{{i}}})(p,q))}}{{\sum\nolimits_{j = 1}^N {\exp (d({F_D},{{{b}}_j})(p,q))} }}
\label{eq1}
\end{equation}

Here, ${{\bf{s}}_i}(p,q)$ represents the similarity map at location $(p,q)$, which measures the softmax-normalized similarity between the defective feature and the texton ${b_i}$, and we adopt the cosine similarity as the distance metric $d({F_D},{b_i})(p,q)$:
\begin{equation}
d\left(F_{D}, b_{i}\right)=\left\langle F_{D}{ }^{*}, b_{i}{ }^{*}\right\rangle=\left\langle\frac{F_{D}}{\left\|F_{D}\right\|}, \frac{b_{i}}{\left\|b_{i}\right\|}\right\rangle 
\label{eq1}
\end{equation}
where $ \left \| \cdot \right \| $ denotes the modulus of a tensor and $\left \langle \cdot \right \rangle $ denotes the inner product, which is elegantly implemented as a convolution operation with stride $K$:
\begin{equation}
\begin{aligned}
\left\langle F_{D}{ }^{*}, b_{i}{ }^{*}\right\rangle(p,q) & =   ({F_D^{*}}*{b_i^{*}})(p,q)\\                    
&= \sum\limits_{m = 0}^{K - 1} {\sum\limits_{n = 0}^{K - 1} {{F_D^{*}}(\eta p + m,\eta q + n){b_i^{*}}(} } m,n)
\end{aligned}
\label{eq1}
\end{equation}
where $*$ denotes the convolution operation and $\eta$ is the scale factor of the downsampling effect of the convolutional layer, which can be calculated as $K$. We further encourage the coherency and flexibility of the similarity map and enrich the gradients during training\cite{r38} by applying the standard convolution operation to ${\bf{s}_i}$ to obtain a new similarity map with ${\bf{s}}_i^*$.
Finally, we reuse the textons ${b_i}$ as deconvolutional filters to reconstruct the output edited feature map, expressed as follows:
\begin{equation}
G = \sum\limits_{i = 1}^N {{{\bf{s}}_i^*} \otimes } {b_i}
\label{eq1}
\end{equation}
where $\otimes$ is the deconvolution operation and $G$ is the output edited feature map, $G \in {R^{{H_F} \times {W_F} \times C}}$.

\begin{figure}[t]
\centerline{\includegraphics[width=\columnwidth]{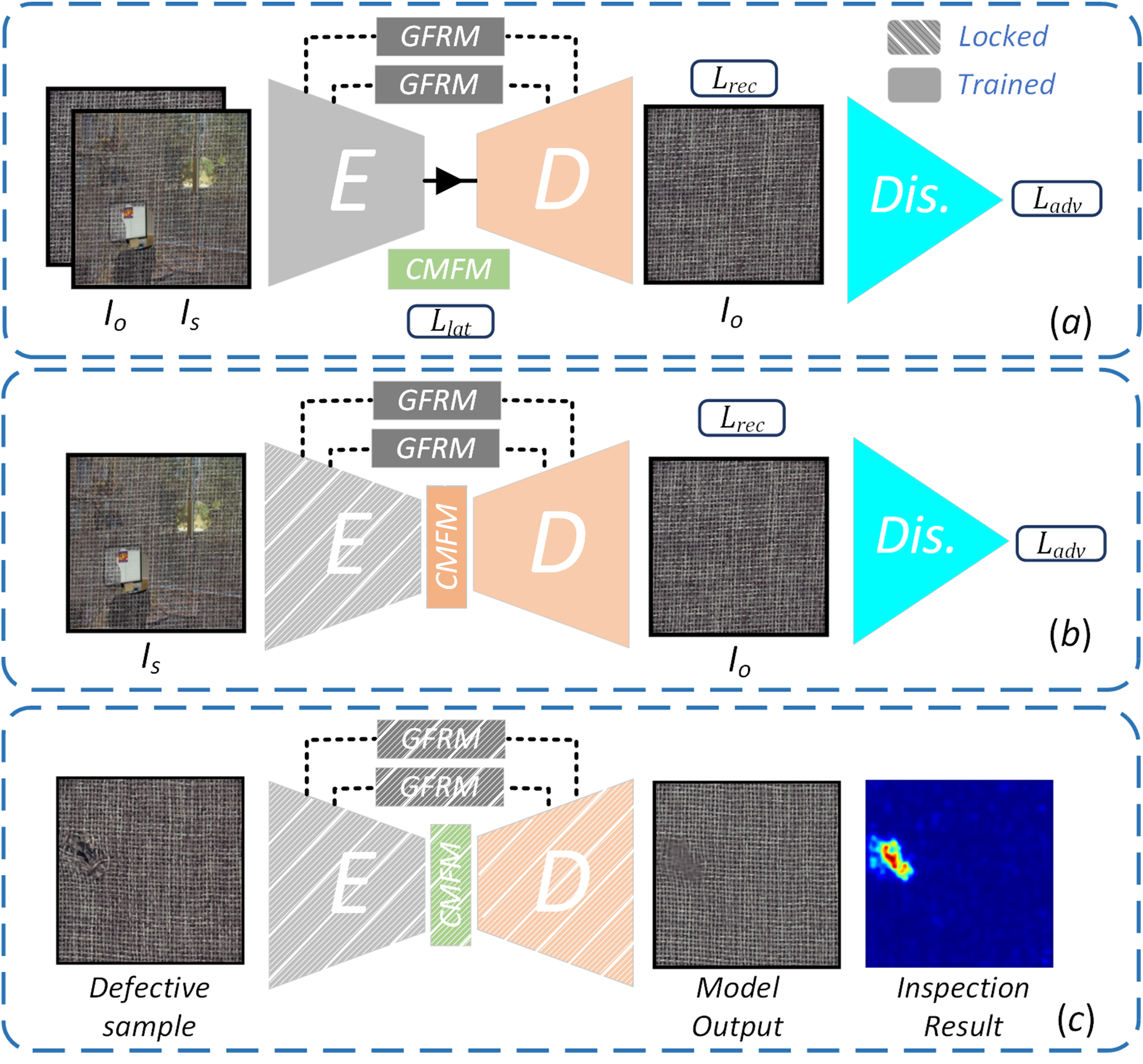}}
\caption{Illustration of the training and testing stages of FMR-Net. (a) Training phase 1. (b) Training phase 2. (c) Testing stage.}
\label{fig8}
\end{figure}

{Fig. \ref{fig7} (a) demonstrates how the GFRMs are embedded onto the different scales of FMR-Net to capture the basic textured microstructures of different receptive fields. In general, as the network level deepens, the feature maps are more semantic, while detail retention becomes progressively worse. To ensure the semantic representation ability of textons, the feature level of each GFRM should not be too shallow; additionally, the low-level feature maps with more precise details also contribute to the accuracy of reconstruction. To obtain a good tradeoff between the semantic representation and the detail maintenance, we deploy the GFRMs in the two deepest layers (${EL_4}$/${DL_4}$, ${EL_3}$/${DL_3}$) of FMR-Net, and the first two shallow layers (${EL_1}$/${DL_1}$, ${EL_2}$/${DL_2}$) are discarded, which can extract semantic primitives with a degree of detail and restrain the model complexity simultaneously.}

{To intuitively validate and visualize the rearrangement efficacy of the GFRM, we present a functional schematic at different levels. As shown in the first row of Fig. \ref{fig7} (b), a real defective image and a defect-free image are used to demonstrate the original feature map, and the memory-generated feature map serves as low-level features in the DNN. The output edited feature map produced in the GFRM is also displayed. Please note the yellow borders and lines highlighted in the figure. The borders indicate where a destructive defect has been edited into a normal texture region, while the lines indicate the specific edge grain existing in the background, which is well preserved even though it is nonexistent in the memory-generated feature map. This phenomenon occurs because in accordance with the `texton principle', all texture patterns are combined with various textons, which means that each background grain can be finely reconstructed through feature rearrangement under the guidance of any texton-retaining normal image. The second row of Fig. \ref{fig7} (b) shows a set of synthetic 5×5 checkerboards utilized for high-level feature simulation, where normal background textures are denoted in blue and defects are denoted in red. The rearrangement results demonstrate that the GFRM achieves the simultaneous goals of background texture retention and defect restoration.}

\subsection{Two-phase training strategy}
For accurate background reconstruction and defect restoration, a novel two-phase training strategy based on a multiple loss function is established to optimize the entire FMR-Net model. The training stage addresses two main learning tasks: background reconstruction and defect restoration.

As shown in Fig. \ref{fig8} (a), for \emph{training phase 1}, the learning tasks are background reconstruction and the establishment of the memory bank in the CMFM. The encoding module (encoder), the GFRM, the decoding module (decoder), the auxiliary classifier ${C_A}$, and the discriminator $D$ are all optimized in this phase. To accurately reconstruct the normal background texture, the optimization process aims to minimize the distance between a defect-free image and the corresponding reconstructed image. Thus, the mean square error is employed as a metric:
\begin{equation}
{L_{rec}}({I_o},I_o^{rec}) = \frac{1}{{{N_o}}}\sum\limits_{{n_o} = 1}^{{N_o}} {{{\left\| {{}^{{n_o}}{I_o} - {}^{{n_o}}I_o^{rec}} \right\|}^2}}  + \varepsilon \sum\limits_{{\bf{\omega }} \in \{ W\} } {{{\left\| {\bf{\omega }} \right\|}_F}} 
\label{eq1}
\end{equation}
where $^{n_o}I$ denotes the ${n_o}$th defect-free image, ${}^{{n_o}}I_o^{rec}$ represents the ${n_o}$th reconstructed image, ${N_o}$ is the batch size, ${\bf{\omega }}$ is the set of weight matrices in FMR-Net, and ${\rm{0 < }}\varepsilon < 1$ is a penalty factor for the regularization term. For further improvement of the reconstruction capability of the generator $G$ (the FMR-Net model), adversarial learning based on the discrepancy between the defect-free image ${I_o}$ and the reconstructed image $I_o^{rec}$ is utilized, where we use the discriminator $D$ to classify the real and fake samples and encourage the perceptual similarity. The adversarial loss is defined as follows:
\begin{equation}
L_{adv}\left(I_{o}, I_{o}^{r e c}\right)=\frac{1}{N_{o}} \sum_{n_{o}}^{N_{o}} [L_{gan}\left(I_{o}, I_{o}^{r e c}\right)+L_{per}\left(I_{o}, I_{o}^{r e c}\right)]
\end{equation}

Here, one objective of adversarial learning is to achieve $\mathop {\min }\limits_G \mathop {\max }\limits_D {L_{gan}}$, which is defined as follows:

\begin{equation}
\begin{aligned}
 L_{gan}\left(I_{o}, I_{o}^{r e c}\right)&=   \underset{I_{o} \sim P_{I_{o}}}{\mathbb{E}} \left[ {\log (1 - D({~}^{{n_o}}I_o^{rec}))} \right]\\ &+ \underset{I_{o} \sim P_{I_{o}}}{\mathbb{E}} \left[ {\log D({~}^{{n_o}}{I_o})} \right]
\end{aligned}
\end{equation}

while we also ensure that the model is capable of generating results that perceptually resemble real samples:
\begin{equation}
L_{per}\left(I_{o}, I_{o}^{r e c}\right)= \sum_{l}^{L} \Psi ^{(l)}\left\|D^{(l)}({}^{{n_o}}I_o)-D^{(l)}({}^{{n_o}}I_o^{rec})\right\|_{1}
\end{equation}
\removelatexerror
\begin{algorithm}[H]
\SetKwInOut{Input}{input}
\SetKwInOut{KwOut}{Parameters}
\caption{Two-phase training strategy}
\label{algorithm}
\KwData{A synthetic data set composed of normal images ${I_o}$ and artificial synthetic images ${I_s}$}
\KwIn{Encoder $\textbf{\emph{E}}$, decoder $\textbf{\emph{D}}$, $\textbf{\emph{GFRM}}$, $\textbf{\emph{CMFM}}$
(auxiliary classifier $\bm{{C_A}}$, addressing network $\bm{{A_N}}$),
discriminator $\bm{Dis}$}

\KwOut{Iterations ${T_1}$ and ${T_2}$, learning rate ${\theta }$}

${\emph{Training phase 1}}$\;

\For{ $i=1,...,{T_1}$}
{
$\left \{ 1 \right \} $.Extract patches ${P^0}$ and ${P^+}$ from ${I_o}$ and patches ${P^-}$ from ${I_s}$\;

$\left \{ 2 \right \} $.Calculate partial derivatives:

${\nabla _E}{L_1}$, ${\nabla _D}{L_1}$, ${\nabla _{\bm{{C_A}}}}{L_{lat}}$, ${\nabla _{GFRM}}{L_1}$, ${\nabla _{Dis}}{L_{adv}}$\;

$\left \{ 3 \right \} $.Superpose partial derivatives \& renew models:

$\bm{E} \leftarrow \bm{E} + \theta  \times {\nabla _E}{L_1}$,
$\bm{D} \leftarrow \bm{D} + \theta  \times {\nabla _D}{L_1}$

$\bm{GFRM} \leftarrow \bm{GFRM} + \theta  \times {\nabla _{GFRM}}{L_1}$

$\bm{{C_A}} \leftarrow \bm{{C_A}} + \theta  \times {\nabla _{\bm{{C_A}}}}{L_{lat}}$ 

$\bm{Dis} \leftarrow \bm{Dis} + \theta  \times {\nabla _{Dis}}{L_{adv}}$\;
}

\SetKwBlock{Evaluate}{Memory bank establishment}{}
\Evaluate{
$\left \{ 1 \right \} $.Extract all patches ${P^+}$ from ${I_o}$\;

$\left \{ 2 \right \} $.Calculate latent representations ${R^+}=E(P^+)$ \;

$\left \{ 3 \right \} $.Execute the \emph{k-means} algorithm on ${R^+}$ to obtain centroids \;

$\left \{ 4 \right \} $.Store cluster centroids as memory bank $\bf{{M}}$\;
}

${\emph{Training phase 2}}$\;

\For{ $i=1,...,{T_2}$}
{
$\bf{\left \{ 1 \right \} }$.Extract patches ${P^0}$ from ${I_o}$ and patches ${P^-}$ from ${I_s}$\;

$\bf{\left \{ 2 \right \} }$.Calculate partial derivatives:

${\nabla _D}{L_2}$, ${\nabla _{\bm{{A_N}}}}{L_{2}}$, ${\nabla _{GFRM}}{L_2}$, ${\nabla _{Dis}}{L_{adv}}$\;

$\bf{\left \{ 3 \right \} }$.Superpose partial derivatives \& renew models:

$\bm{D} \leftarrow \bm{D} + \theta  \times {\nabla _E}{L_2}$,$\bm{{A_N}} \leftarrow \bm{{A_N}} + \theta  \times {\nabla _{\bm{{A_N}}}}{L_{2}}$ 

$\bm{GFRM} \leftarrow \bm{GFRM} + \theta  \times {\nabla _{GFRM}}{L_2}$

$\bm{Dis} \leftarrow \bm{Dis} + \theta  \times {\nabla _{Dis}}{L_{adv}}$\;
}

\end{algorithm}
where $D^{(l)}$ is the $l$th intermediate feature map of $D$ given inputs ${}^{{n_o}}I_o$ and ${}^{{n_o}}I_o^{rec}$ and $\Psi ^{(l)}$ denotes the relative importance of layer $l$. We use the feature maps from the ReLU layers of the 3rd and 4th blocks at each scale and set $\Psi ^{(l)}=1/2$.

As mentioned in the description of the CMFM above, the distribution of the latent features is constrained to improve its discriminability and sparsity. We also introduce the following latent constraint loss function:
\begin{equation}
{L_{lat}}({I_o},{I_s}) =  \frac{1}{{N_o}}\sum\limits_{n = 0}^{{N_o} - 1} {L_{lat}^{hid}({R^0},{R^ + },{R^ - })} 
\end{equation}
Combining (10), (11) and (14), we obtain the loss function for \emph{training phase 1} as follows:
\begin{equation}
\begin{aligned}
{L_1}({I_o},{I_s},I_o^{rec}) = \omega _{rec}^1{L_{rec}}({I_o},I_o^{rec}) &+ \omega _{adv}^1{L_{adv}}({I_o},I_o^{rec}) \\
&+ \omega _{lat}^1{L_{lat}}({I_o},{I_s})
\end{aligned}
\end{equation}
where $\omega _{rec}^1$, $\omega _{adv}^1$, and $\omega _{lat}^1$ are the weights that balance the relative contributions of the three types of losses. We recommend setting $\omega _{rec}^1=100$, $\omega _{adv}^1=1$, and $\omega _{lat}^1=1$ in this paper. After the completion of \emph{training phase 1}, all nondefective images ${I_o}$ are passed through the preliminarily trained encoder to calculate the corresponding latent representations. Then, standard \emph{K-means} clustering is conducted, and the centroids are stored to establish the memory bank. Thus, the feature-memory-based reasoning procedure is initialized, and subsequently, the training of the entire FMR-Net (including the CMFM) will be boosted in \emph{training phase 2}.

As shown in Fig. \ref{fig8} (b), for \emph{training phase 2}, the learning task is defect restoration, and the decoding module (decoder), the addressing network ${A_N}$,the GFRM, and the discriminator $D$ are involved in the optimization process. To eliminate defects and convert them into background texture, an artificial defect image ${I_s}$ is used as the input to FMR-Net, and the source defect-free image ${I_o}$ is naturally utilized as the corresponding generation target. Notably, because the memory bank is introduced in this phase, the parameters of the memory entries, which represent the typical information of normal textures given the specific state of the encoding module (encoder) $E$, are fixed after \emph{training phase 1}. That is, $E$ should be locked to guarantee the correspondence between $E$ and the memory bank.
The goal of the loss function in this phase is to measure the success of defect transformation. Therefore, the loss function for \emph{training phase 2} is as follows:
\begin{equation}
{L_{\rm{2}}}({I_s},I_s^{rec}) = \omega _{rec}^2{L_{rec}}({I_s},I_s^{rec}) + \omega _{adc}^2{L_{adv}}({I_s},I_s^{rec})
\end{equation}

where $I_s^{rec}$ denotes the reconstruction result for the artificial defect image ${I_s}$ and $\omega _{rec}^2$ and $\omega _{adv}^2$ are the corresponding weights. In this paper, $\omega _{rec}^2$ is set to 100, and $\omega _{adv}^2$ is set to 1. We summarize the whole two-phase training procedure in Algorithm 1.

\subsection{Multimodal inspection method}

After training, the FMR-Net model can be employed for textural defect detection. As shown in Fig. \ref{fig8} (c), the encoding module (encoder), the GFRM, the CMFM (memory bank and addressing network), and the decoding module (decoder) are used in this stage, while the discriminator and auxiliary classifier are discarded. To reduce the occurrence of pseudo defects and improve the inspection accuracy, a multimodal inspection method is constructed to obtain the final inspection result.

For reconstruction-based model inference, the per-pixel subtraction operation is commonly used; however, this unimodal method often introduces false detections due to noise corruption and causes misinspection because the specific indicator may not be sensitive to defects. Furthermore, naïve image subtraction assumes that each pixel is independent of its neighboring pixels, ignoring structural information. For this reason, we should instead exploit the metrics with a larger receptive field, which focus on the local structural difference properties of an image.

The gradient magnitude similarity (GMS)\cite{r40} and the structural similarity index (SSIM)\cite{r41} are two well-established metrics that are widely used in the signal processing field. In this paper, both the SSIM and GMS are utilized as local structural similarity metrics. The GMS judges the image distance on the basis of the gradient response, as expressed by the
following equation:
\begin{equation}
GMS(I,{I^{rec}}) = \frac{{2g(I)g({I^{rec}}) + {C_0}}}{{g{{(I)}^2} + g{{({I^{rec}})}^2} + {C_0}}}
\end{equation}
where $I$ and $I^{rec}$ denote the candidate image and the corresponding reconstructed image, respectively; ${C_0}$ is a constant ensuring numerical stability; and ${g(X)}$ is the gradient magnitude map of image $X$:
\begin{equation}
g(X) = \sqrt {{{(X*{p_x})}^2} + {{(X*{p_y})}^2}}
\end{equation}
where ${p_x}$ and ${p_y}$ denote 3×3 Prewitt filters along the x- and y-axes and $*$ represents the convolution operation. Accordingly, we define the modality 1 anomaly map as follows:
\begin{equation}
{A_{{\rm{m1}}}}(I,{I^{rec}}){\rm{ = }}1 - GMS(I,{I^{rec}})
\end{equation}

The SSIM is a popular reference-based index for evaluating local structural changes in images because it is sensitive to the characteristics of the human visual system (HSV)\cite{r41}. The SSIM can be calculated as per the following equation:
\begin{equation}
SSIM(I,{I^{rec}}) = \frac{{(2{\mu _I}{\mu _{{I^{rec}}}} + {C_1})(2{\sigma _I}{\sigma _{{I^{rec}}}} + {C_2})}}{{(\mu _I^2 + \mu _{{I^{rec}}}^2 + {C_1})(\sigma _I^2 + \sigma _{{I^{rec}}}^2 + {C_2})}}
\end{equation}
where the $\mu$ and $\sigma$ notations represent the means and standard deviations of the corresponding images, which are computed using Gaussian filters, and ${C_1}$ and ${C_2}$ are constants ensuring numerical stability. The SSIM values are computed at the patch level using the sliding window strategy described in \cite{r41}. Accordingly, we define the modality 2 anomaly map as follows:
\begin{equation}
{A_{{\rm{m2}}}}(I,{I^{rec}}){\rm{ = }}1 - SSIM(I,{I^{rec}})
\end{equation}

In addition, the per-pixel residual map constitutes the modality 3 anomaly map:
\begin{equation}
{A_{{\rm{m3}}}}(I,{I^{rec}}){\rm{ = }}RM(I,{I^{rec}}) = {\left\| {I - {I^{rec}}} \right\|}
\end{equation}

Finally, the final multimodal inspection result is obtained by fusing the above anomaly maps using the dot product operation, as shown in the following equation:
\begin{equation}
A(I,{I^{rec}}) = {A_{{\rm{m1}}}}(I,{I^{rec}}) \cap {A_{{\rm{m2}}}}(I,{I^{rec}}) \cap {A_{{\rm{m3}}}}(I,{I^{rec}})
\end{equation}

In practical situations, a median filter operation can be applied to the anomaly image of each modality to obtain smoother results. This multimodal image fusion strategy enables the proposed FMR-Net method to achieve more robust and accurate performance.

\begin{table*}
\caption{Detailed configuration of the data set}
\label{table}
\setlength{\tabcolsep}{3pt}
\begin{tabular}{p{\textwidth}}
$\includegraphics[width=\textwidth]{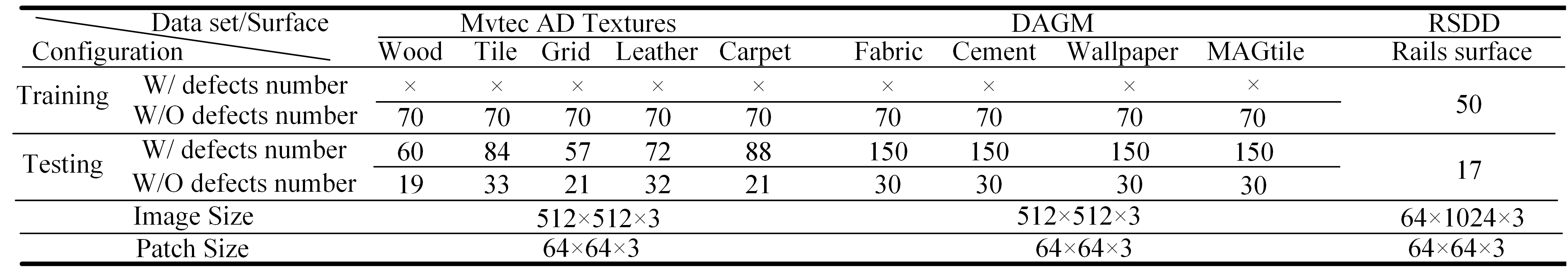}$
\end{tabular}
\label{table1}
\end{table*}

\section{Performance Verification Experiments}
In this section, we report a set of experiments conducted to
evaluate the performance of the proposed FMR-Net method.
Specifically, to investigate the influence of each component of FMR-Net, including the two-phase training strategy, the CMFM, the GFRM and the multimodal inspection method, a series of ablation analyses are presented. Then, the overall comparison experiments are conducted to qualitatively and quantitatively compare the inspection performance of FMR-Net with that of several state-of-the-art methods. Furthermore, we propose a multilevel detection method for fast inference in scenarios of defect/background imbalance, which enables the practical deployment of FMR-Net in real industrial applications of edge-cloud collaborative intelligent manufacturing.

\subsection{Quantitative evaluation criterion}
To quantitatively evaluate the performance of various methods, we adopt the AUC ROC, precision, recall and $F1$-measure as the verification indicators. The AUC ROC refers to the area under the receiver operating characteristic (ROC) curve and reflects the classification ability of a model on the positive and negative examples; thus, it has been commonly used for inspection accuracy verification in the mainstream related work and benchmarks of anomaly detection. However, the AUC ROC is independent of the absolute value of the threshold and may not reflect the accuracy of the assay under the specific threshold situation. Therefore, the precision, recall and $F1$-measure, which reflect the comprehensive evaluation of these two metrics, are also utilized for a more objective assessment.
\subsection{Datasets and implementation details}
In these experiments, a comprehensive dataset composed of various textural samples was utilized, including homogeneous and nonregularly textured surfaces. Among these data, the carpet, leather, grid, tile, and wood samples were sourced from the MVTec AD texture dataset\cite{r43}; the fabric, cement, wallpaper and MAGtile samples were taken from the DAGM dataset\cite{r44}; and the rail surface images were selected from the RSDDs dataset\cite{r45}. All of these samples are representative of real-world problems and have been widely used for performance verification. All images in the MVTec AD textures dataset and the DAGM dataset were resized to 512×512 pixels, and the slender images in the RSDDs dataset were resized to 1024×64 pixels. Considering the difficulty of reconstructing the whole original-size images, the raw images were sliced into 64×64-pixel patches using the sliding window approach. The detailed configuration of the dataset is reported in Table \ref{table1}. Please note that in the RSDDs dataset, all of the raw images are defective, and we included only defect-free patches in the training set for model training.

All experiments were implemented on a computer with a Xeon(R) Silver 4110 processor and 64 GB of RAM, which was equipped with four NVIDIA 2080Ti GPUs and the Ubuntu 16.04 operating system. The whole FMR-Net was optimized using the Adam optimizer with a learning rate of 0.001. The iteration sizes for the two-phase model training process were set to 200 000 and 100 000.

\begin{table}
\caption{Ablation study of FMR-Net on the wood data set}
\label{table}
\setlength{\tabcolsep}{3pt}
\begin{tabular}{p{\columnwidth}}
$\includegraphics[width=\columnwidth]{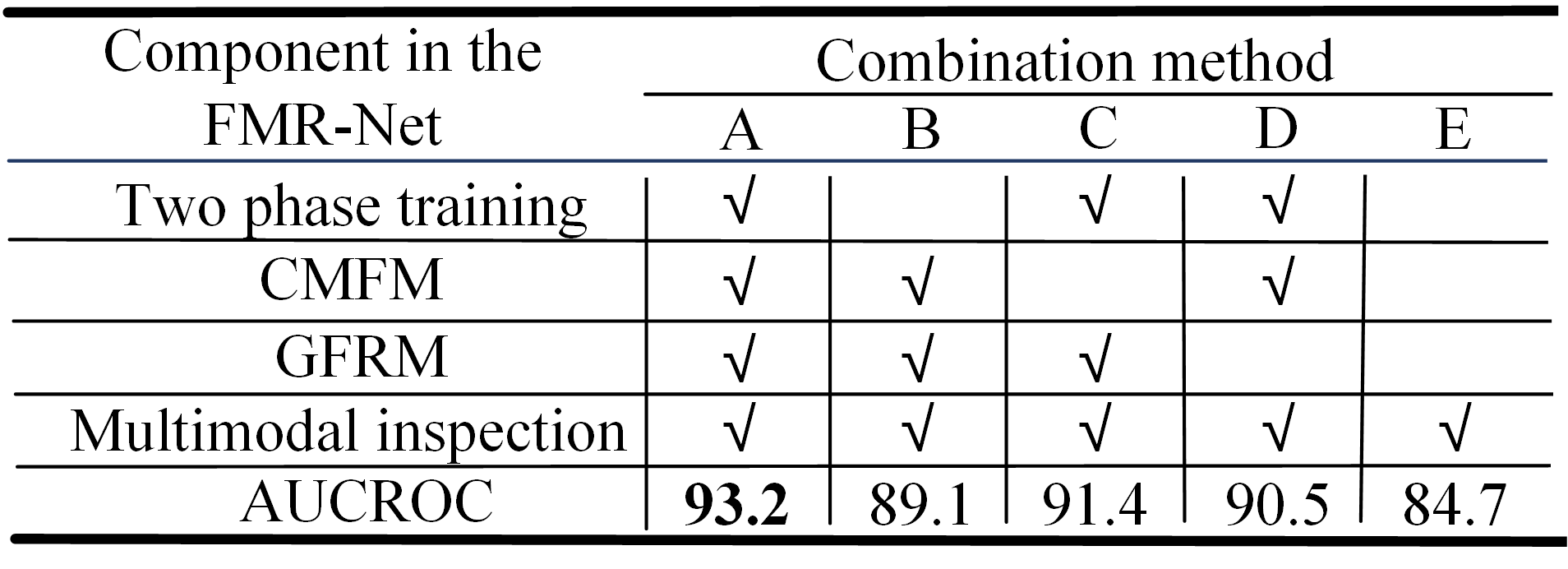}$
\end{tabular}
\label{table2}

\end{table}

\begin{figure}[t]
\centerline{\includegraphics[width=\columnwidth]{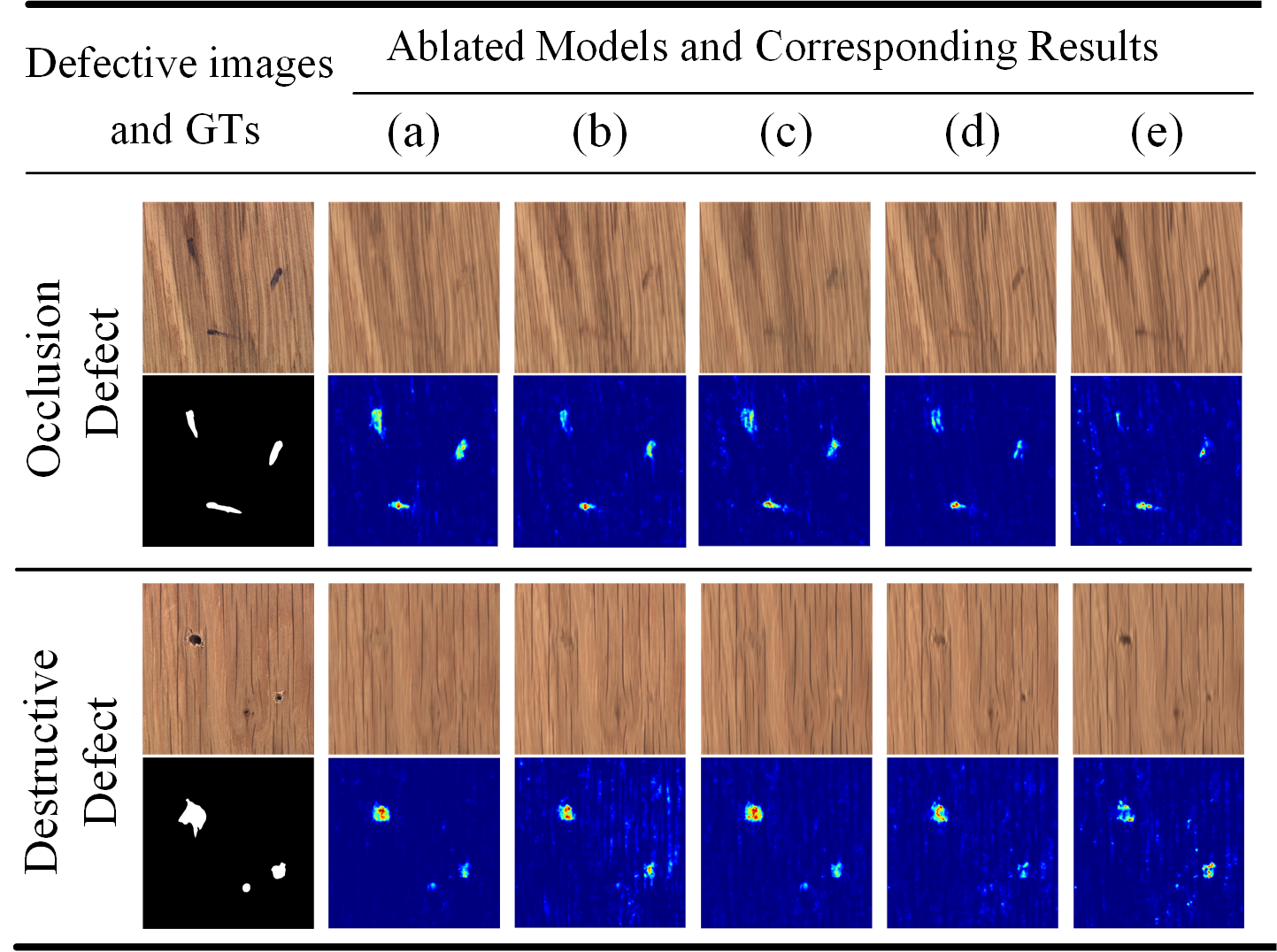}}
\caption{Examples of two typical images from tests performed in the ablation studies. The first row displays the reconstructed images, and the second row displays the inspection results. Each column corresponds to the configuration in Table II.}
\label{fig9}
\end{figure}

\begin{table}[t]
\caption{
AUC ROC comparison of the inspection performance achieved with different modal combinations on the tile data set
}
\label{table}
\setlength{\tabcolsep}{3pt}
\begin{tabular}{p{\columnwidth}}
$\includegraphics[width=\columnwidth]{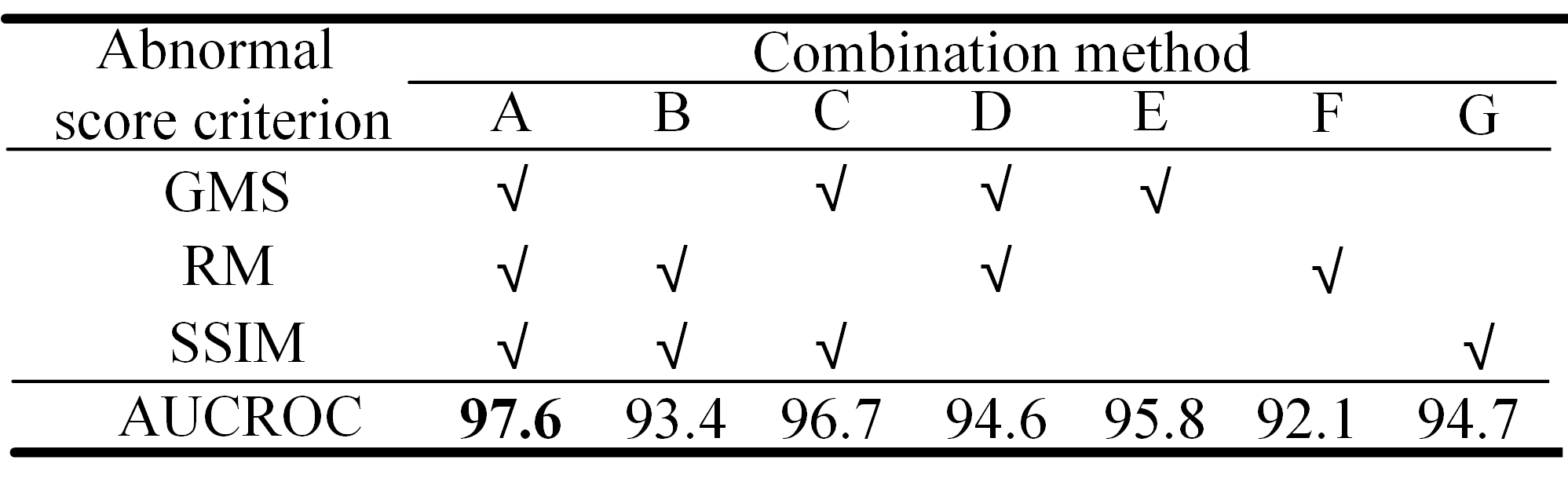}$
\end{tabular}
\label{table3}
\end{table}

\begin{figure}[t]
\centerline{\includegraphics[width=\columnwidth]{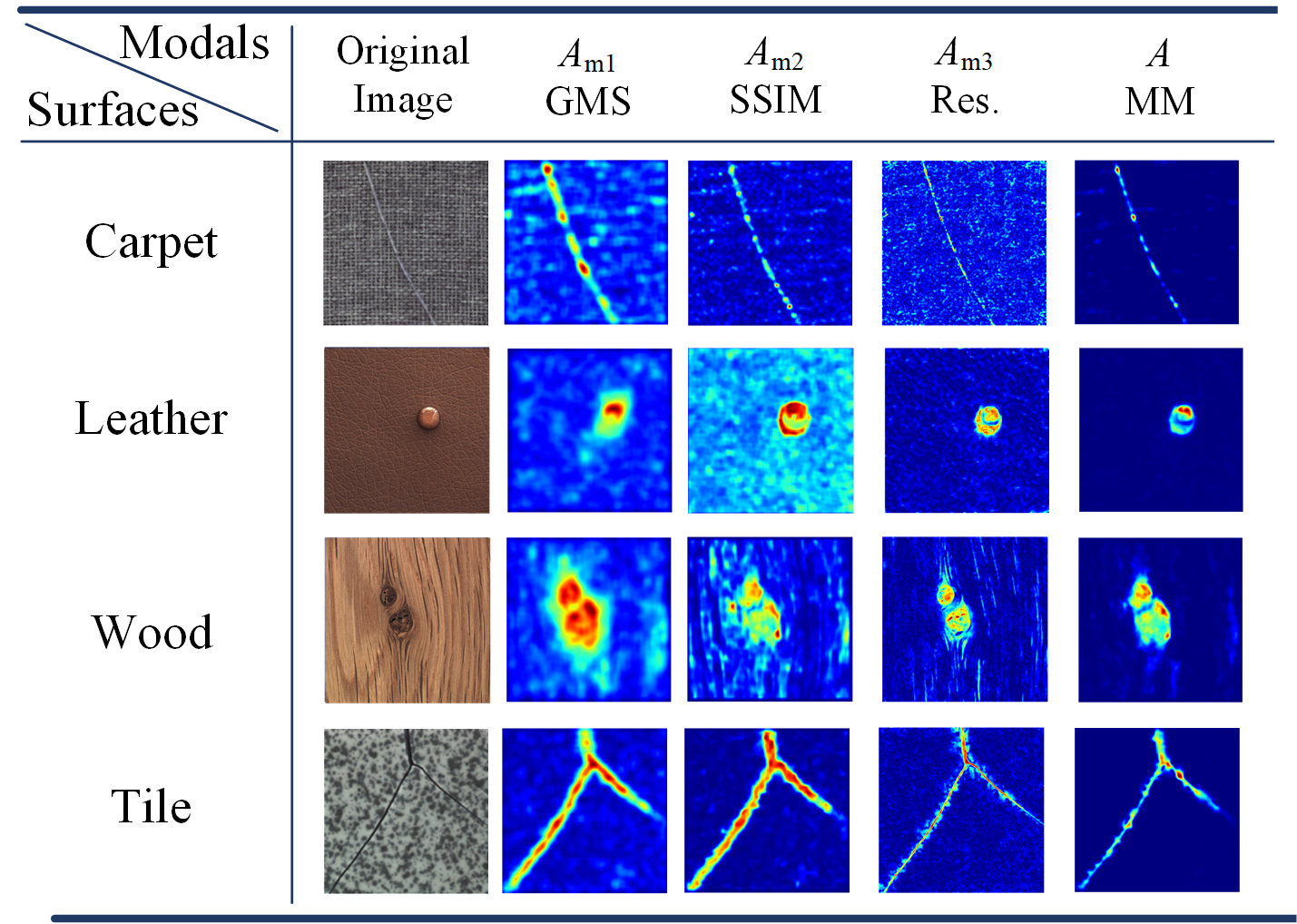}}
\caption{
Localizing texture defects with the proposed multimodal inspection method on carpet samples (first row), leather samples (second row), wood samples (third row) and tile samples (fourth row). From left to right are shown the original defect images, the anomaly maps for each independent modality, and the multimodal inspection results.}
\label{fig10}
\end{figure}

\subsection{Ablation studies of FMR-Net}

A set of ablation studies was conducted to analyze the effectiveness of each component of FMR-Net. Notably, the parameter settings were kept the same during these experiments to ensure fair comparisons. The results of the ablated variant models on selected defective samples are shown in Fig. \ref{fig9}, and the AUC ROC results for each model are presented in Table \ref{table2}.

\subsubsection{Necessity of the two-phase training strategy}
In FMR-Net, the two-phase training strategy is a novel technique that enables the model to recognize anomalies on the basis of extra artificial synthetic defect images. The model receives a synthetic defect image ${I_s}$ as input and utilizes the corresponding original defect-free image ${I_o}$ as its target. Thus, the model gains the ability to restore real defects. The experimental results obtained without the two-phase training strategy are described in column B of Table \ref{table2}. When the entire FMR-Net model is trained on only defect-free images, the AUC ROC value decreases by 4.1 compared to that of the fully trained model (column A). Additionally, the results for two typical types of defective examples are shown in column (b) of Fig. \ref{fig9}, from which we can see that residual defects are still present.

These experimental results reveal that the two-phase training strategy can benefit the overall performance. Introducing extra artificial synthetic defective images into the training process improves the defect restoration ability, which is beneficial for the elimination of defects.

\subsubsection{Influence of the CMFM} To improve the discriminability of the latent representations, we introduce the CMFM to enhance the semantic extraction capability of the encoder by means of self-supervised learning. Without the CMFM, the latent representations remain entangled, which further influences the GFRM.

Column C of Table \ref{table2} lists the experimental results. The CMFM(column A) improves the AUC ROC by a margin of 1.8 compared to the model without the CMFM. In addition, as shown in Fig. \ref{fig9} (c), without the CMFM, the reconstruction of the background and the restoration of the defects are both inefficient. This is because when the corresponding features are entangled with each other, their representative capability is limited. This loss of information leads to inferior background textural reconstruction, hence causing noise corruption. Moreover, without the CMFM, the unstable normality of the memory-generated feature map is unable to guarantee the normal texture semantics of the textons in the GFRM, thus resulting in poor restoration ability of FMR-Net. As seen from the results for the destructive defect in Fig. \ref{fig9} (c), although the hole is exchanged for other grains, the naturalness is worse than that in Fig. \ref{fig9} (a).

\begin{figure}[t]
\centerline{\includegraphics[width=\columnwidth]{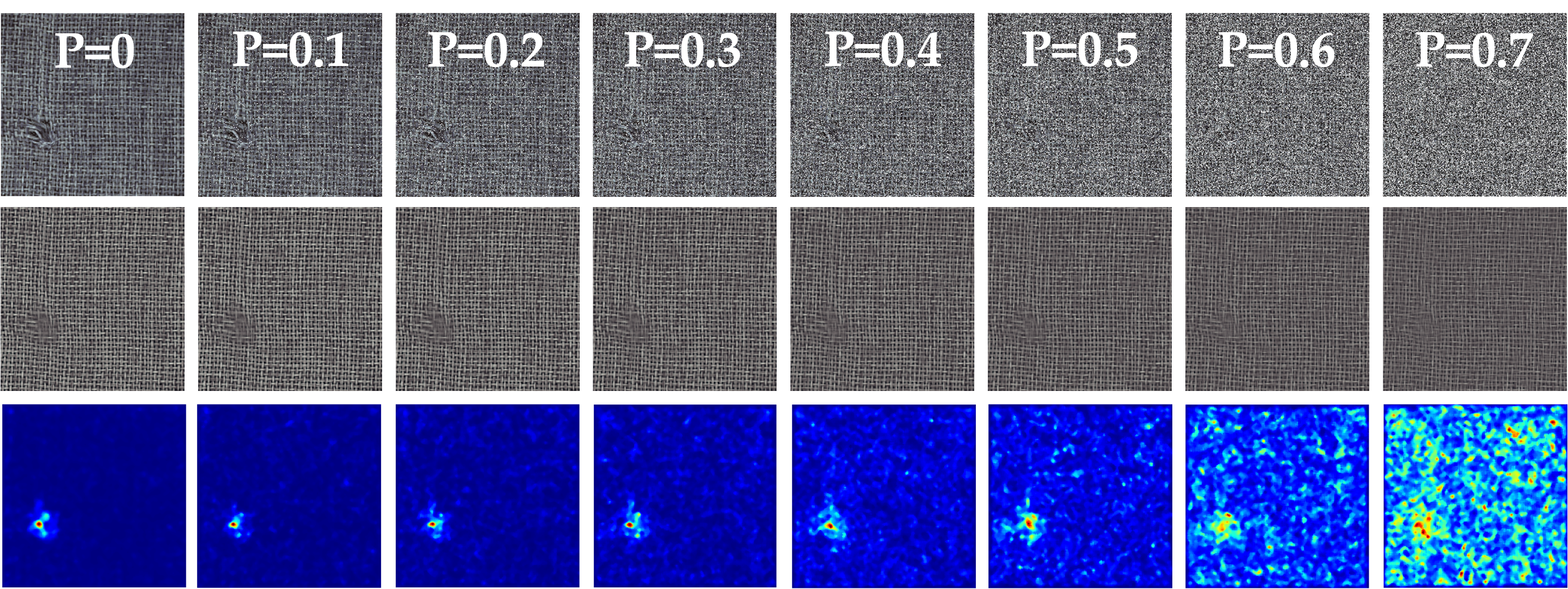}}
\caption{
Visualization results of the robustness of the detection performance against noise corruption. First row: noise-contaminated samples. Second row: reconstruction results. Third row: inspection results.}
\label{fig11}
\end{figure}

\begin{figure}[t]
\centerline{\includegraphics[width=\columnwidth]{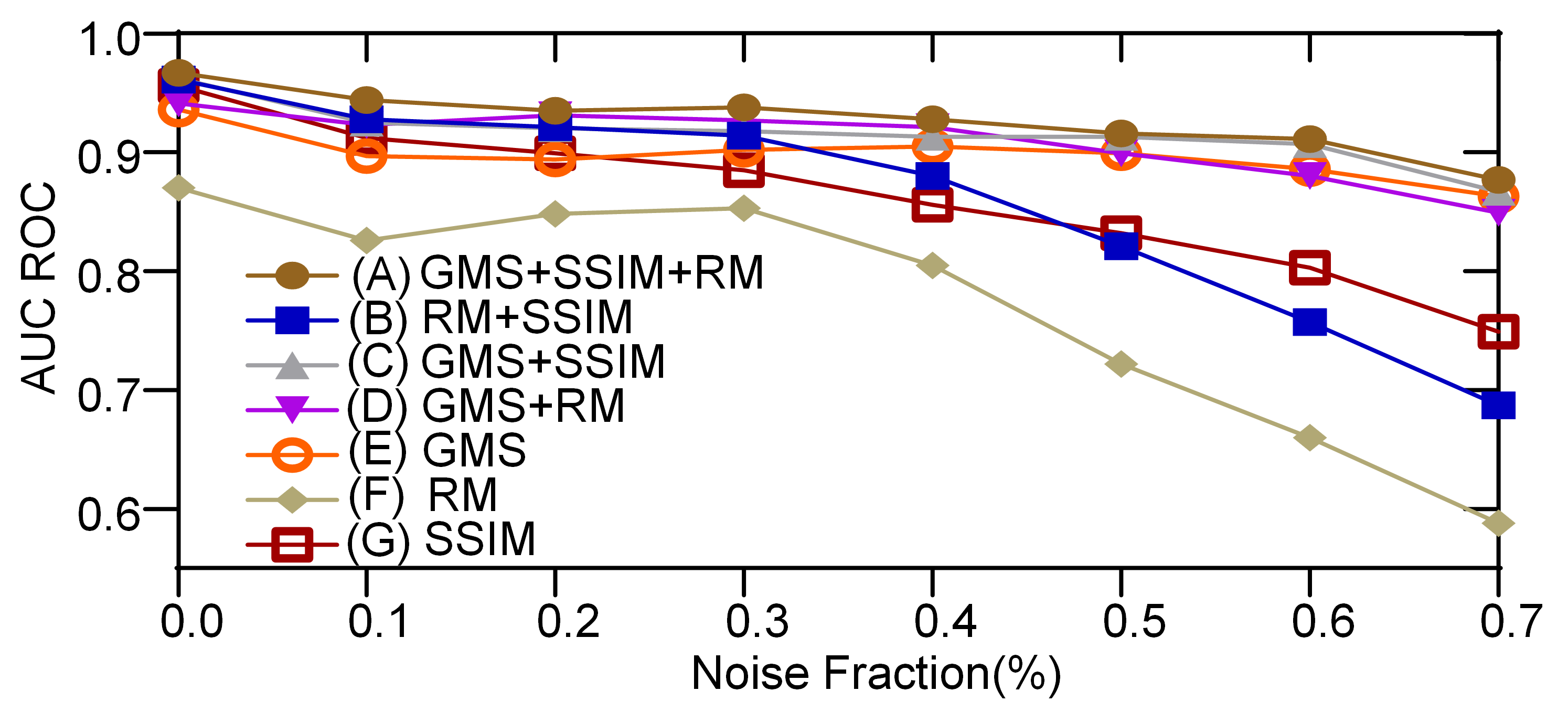}}
\caption{
Overall AUC ROC inspection performance against different proportions of noise for model testing.}
\label{fig12}
\end{figure}

These experiments demonstrate that the CMFM can improve inspection performance due to the discriminative-feature-constrained encoder optimization and the feature-memory-based reasoning procedure, both of which improve the accuracy of background reconstruction and the naturalness of defect restoration.

\begin{figure}[t]
\centerline{\includegraphics[width=\columnwidth]{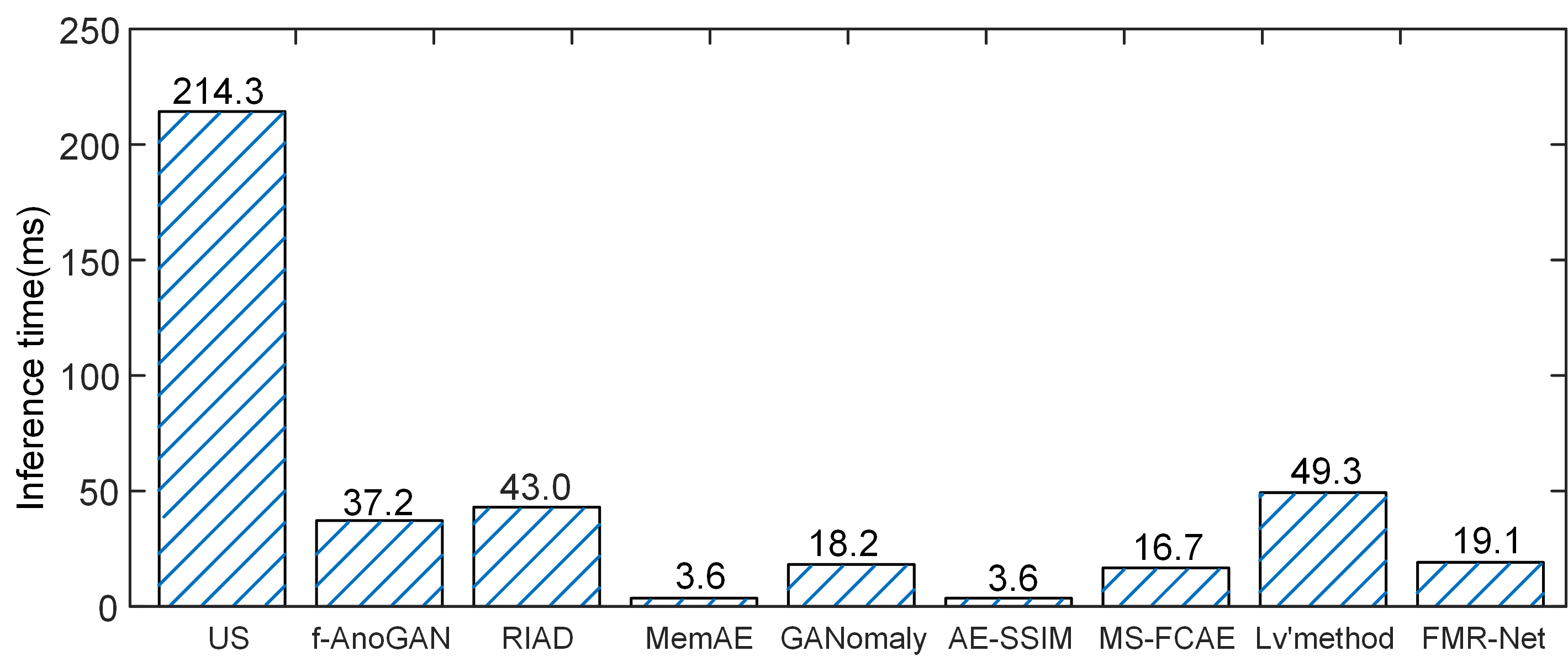}}
\caption{
Comparison of time consumption during inference.}
\label{fig13}
\end{figure}

\subsubsection{Effect of the GFRM} The GFRM is proposed to rearrange the features on the skip connection pathways to suppress the reconstruction of residual anomalous features on the basis of texton theory. As described in Section III-D, the GFRM plays a crucial role in removing persistent defects.

Column (d) of Fig. \ref{fig9} shows an example of the results of FMR-Net without the GFRM. Please specifically note the results for defects of the destructive type. In the absence of the GFRM, FMR-Net is still capable of processing defects with dramatic variations in pixel values, such as hole regions, but the textural patterns remain anomalous, as illustrated by the vague residual hole patterns remaining in the background. By comparison, the results in column (a) show that such persistent destructive defects are removed from the background, thus improving the inspection performance by a margin of 2.7, as reported in Table \ref{table2}.

This experiment verifies the capacity of the GFRM for persistent defect suppression, which arises from its ability to rearrange defects into background textures while maintaining elaborate texture details due to multiscale feature parsing.

When all three of the above modules are removed, the model can be viewed as a basic GAN composed of an encoder, a decoder and a discriminator. An example of the inspection results is shown in column (e) of Fig. \ref{fig9}. The basic GAN model reconstructs the defect with high fidelity during the testing phase, which leads to misinspection and overinspection. Table \ref{table2} shows that the AUC ROC in column E is decreased by 8.5 compared to that of FMR-Net (column A).

\begin{table*}
\caption{AUC ROC results of different methods on five types of textures in MVTec AD}
\label{table}
\setlength{\tabcolsep}{3pt}
\begin{threeparttable}
\begin{tabular}{p{\textwidth}}
$\includegraphics[width=\textwidth]{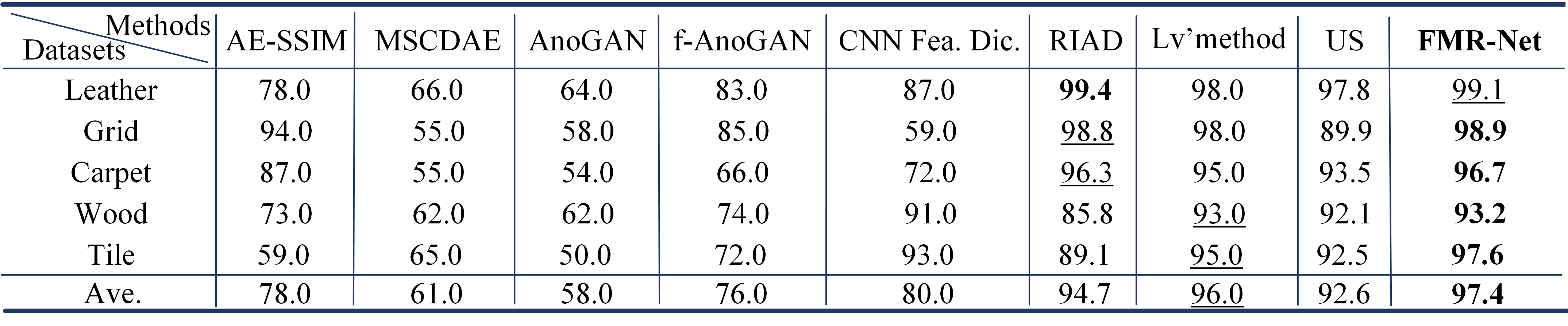}$
\end{tabular}
\begin{tablenotes}
       \footnotesize
       \item[1]The best AUC ROC performance is indicated by bold font, while the second best is indicated by an underline.
\end{tablenotes}
\end{threeparttable}
\label{table4}
\end{table*}

\begin{figure*}[t]
\centerline{\includegraphics[width=\textwidth]{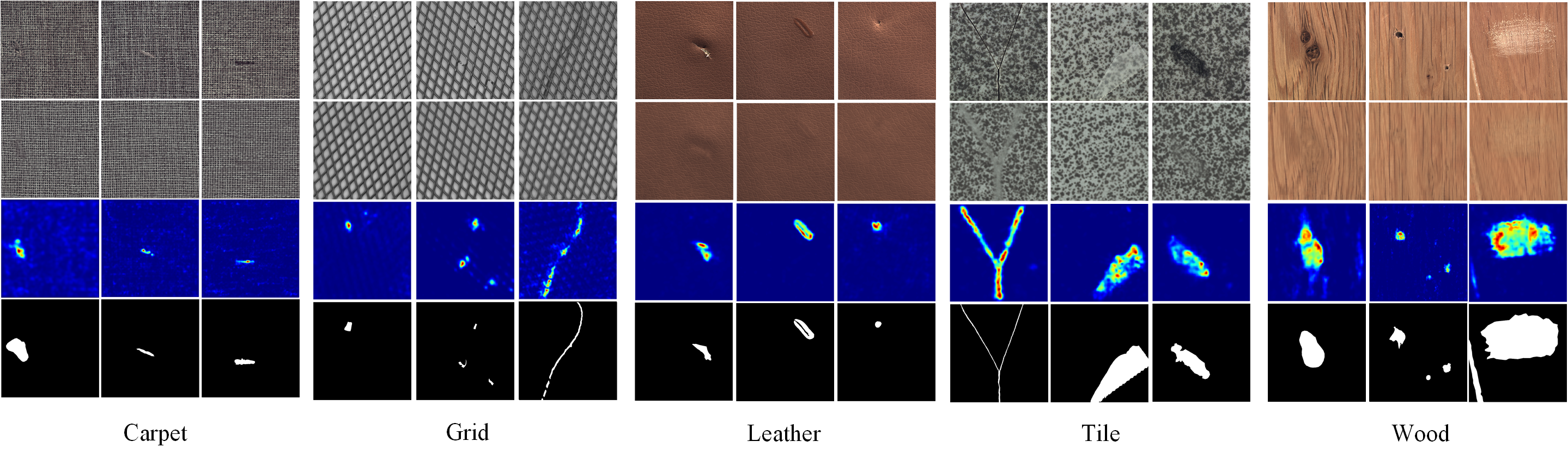}}
\caption{
Reconstruction and detection results for various textural samples. From top to bottom are shown the original defect samples, the reconstruction results, the inspection results obtained using the proposed FMR-Net model, and the ground truth.
}
\label{fig14}
\end{figure*}

\begin{figure*}[t]
\centerline{\includegraphics[width=\textwidth]{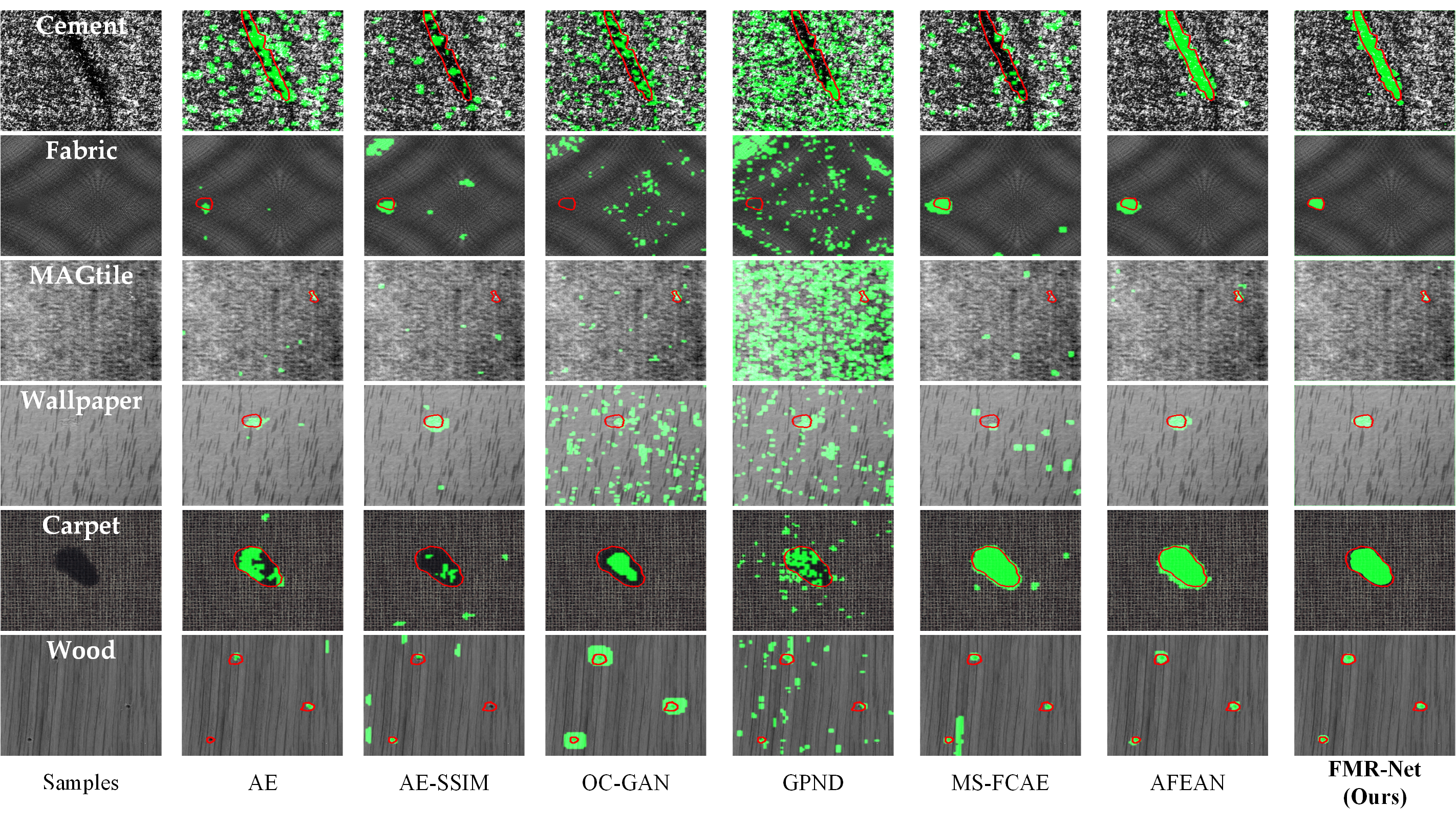}}
\caption{
Examples of the defect inspection performance of the compared methods for the defect binarization detection task, in which the ground truth is identified by the red contours, and the segmented results are indicated in the green area.}
\label{fig15}
\end{figure*}

\begin{table*}
\caption{Precision,recall and $F1$-measure results for different methods on six types of textures}
\label{table}
\setlength{\tabcolsep}{3pt}
\begin{tabular}{p{\textwidth}}
\centerline{$\includegraphics[width=154mm]{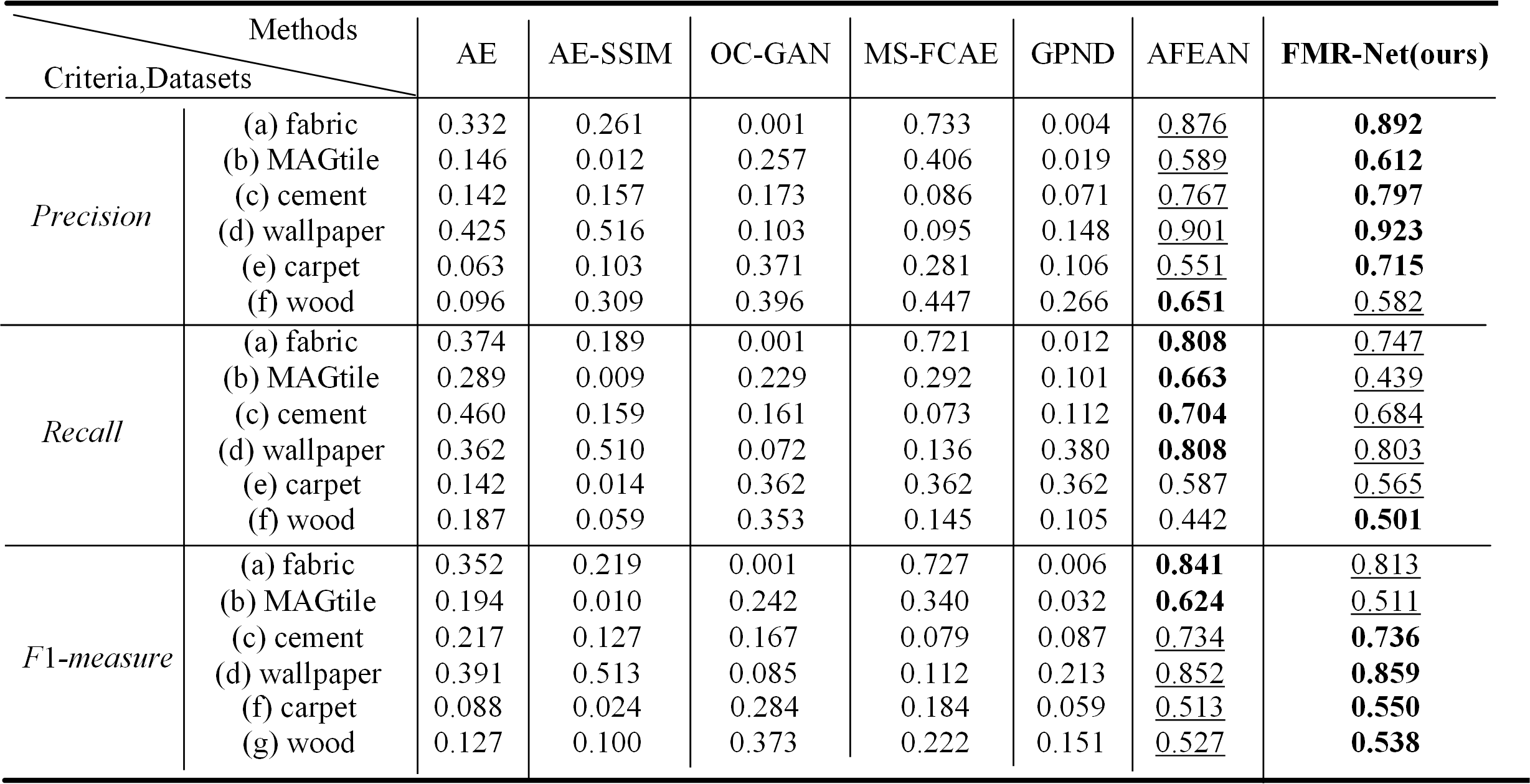}$}
\end{tabular}
\label{table5}
\end{table*}

\subsubsection{Improvements due to the multimodal inspection method} To illustrate the effectiveness of the multimodal inspection method, we experimented with FMR-Net models to test various anomaly score criteria.

Fig. \ref{fig10} shows the three anomaly maps of the different independent modalities and the multimodal (MM) anomaly maps for several types of texture surfaces. The motivations for these experiments are summarized as follows: (1) Different textural surfaces or defects are likely concealed or manifested in different modalities. As shown in Fig. \ref{fig10}, different surfaces will present different activation responses in the same modality, and thus, each has its own optimal inspection modality. Selecting the most suitable modality in advance demands prior knowledge and is impractical. (2) The unimodal methods suffer from noise corruption, as seen in the first three columns of Fig. \ref{fig10}. (3) Defective regions tend to exhibit high responses in different modalities, while noise manifests only in a particular modality.

To leverage hybrid multimodal efficacy, we propose a novel multimodal inspection method, as shown in the rightmost column of Fig. \ref{fig10}. Through contrastive analysis of the fused results (the rightmost column) and the results from the independent modalities (first three columns), the conclusion can be drawn that the multimodal mechanism is capable of facilitating robustness and eliminating false alarms. Many falsely detected defective regions are inevitable in a unimodal approach, e.g., in ${A_{m1}}$ for carpet and in ${A_{m2}}$ for wood, but the fused results are accurate. Similarly, although misinspection occurs in some cases, as in the ${A_{m3}}$ results for carpet and wood, the final results are still satisfactory. The evaluation results for all possible modal combinations are listed in Table III. It can be observed that the multimodal inspection method is superior in all cases.

Therefore, the proposed FMR-Net model tends to be more robust and accurate with multimodal inspection.

\subsection{Inspection robustness against the noise fraction}
An industrial inspection model must be equipped with robustness under uncertainties. To further investigate the stability properties of the model, we experimented with FMR-Net models by testing various proportional noise corruption. Probabilistic speckle noise is utilized for image corruption in experiments; that is, the pixel intensities of the images will be replaced with uniformly distributed random values by the probability \emph{p}.

As illustrated in Fig. \ref{fig11}, the phenomenon of texture degradation becomes progressively worse as the noise probability increases, yet the reconstruction results remain stable. This phenomenon occurs because the artificial synthetic defect in the two-phase training strategy enables the model to be noise resistant, and the normal texture prototypes in the CMFM and the GFRM also guarantee the reliability of the reconstruction results.

Furthermore, as shown in the third row of Fig. \ref{fig11}, the defect location results are still favorable even under heavy noise contamination, which is attributed to the proposed multimodal inspection method. As the additional results exhibited in Fig. \ref{fig12}, the AUC ROC performance of several modal combinations decreases greatly, while the multimodal method remains relatively robust against the increasing noise fraction.

\subsection{Overall Comparative Experiment Results and Analysis}

To verify the performance of the proposed FMR-Net method, in this section, the inspection performance of FMR-Net is compared with that of several existing state-of-the-art methods in specific industrial scenarios.

\subsubsection{Comparative results for processing efficiency}
A practical model needs to meet the real-time inspection demands. To verify the detection efficiency of the proposed method, the time consumption of the proposed FMR-Net is compared with several benchmarks in this subsection. These include AE-SSIM\cite{r46}, MemAE\cite{r23}, f-NoGAN\cite{r47}, GANomaly\cite{r37}, RIAD\cite{r24}, MS-FCAE\cite{r9}, Lv's method\cite{r28}, and Uninformed Students (US) method\cite{r48}. For each technique, the average inference time (AIT) was measured for 512 × 512 image sizes under the same hardware conditions.

The quantitative comparison results are presented in Fig. \ref{fig15}. The AIT of the FMR-Net was 19.1, which was ranked behind the AE-SSIM, MemAE, GANomaly and MS-FACE. However, FMR-Net performed much better than the above methods in terms of detection performance. For models with relatively preferable inspection capabilities, such as RIAD, Lv's method and US, more computational time consumption is needed, especially for the US method, which limits their practical application effectiveness.

This experiment reveals that the proposed FMR-Net can achieve accurate detection results with acceptable processing efficiency and is able to meet high-speed industrial inspection demands.

\subsubsection{Comparative results for the defect anomaly detection task}
Following previous work\cite{r43}, first, the following competing approaches are evaluated on the MVTec AD dataset for performance comparison of anomaly detection: AE-SSIM\cite{r46}, MSCDAE\cite{r17}, AnoGAN\cite{r18}, f-AnoGAN\cite{r47}, the CNN feature dictionary method\cite{r37}, RIAD\cite{r24}, Lv's method\cite{r28}, the Uninformed Students (US) method\cite{r48} and the proposed FMR-Net.

The quantitative comparative results in terms of the AUC ROC indicator are given in Table \ref{table4}. AE-SSIM shows superior performance on simple homogeneous textures such as grids. However, it is incapable of inspecting irregular textures such as tiles and wood. MSCDAE does not perform well due to the lack of semantic information gained due to the small receptive field. Early AE-based models can easily reconstruct defects as well as background; thus, the remaining anomalous patterns influence the subsequent detection process. AnoGAN still achieves unsatisfactory results, and our experiments reveal that its ability to reconstruct normal textures is limited. This can be attributed to the fact that the unstable latent representation search causes the quality of the generated images to be compromised. In f-AnoGAN, an extra encoder is introduced to make it more robust for background reconstruction, leading to better and faster performance. However, the experimental results illustrate that it is powerless in repairing destructive defects, for which anomalous patterns will remain in the reconstructed images. For the feature-domain detection method known as the CNN feature dictionary method, although it achieves reasonable results on irregular textures such as wood and tiles, it is unable to handle all situations and categories well simultaneously. RIAD achieves excellent defect elimination and shows superior performance on homogeneous surfaces such as leather, grids and carpet. Unfortunately, this method leads to a poor normal texture reconstruction effect for textures with strong randomness, such as tiles and wood. Thus, the nonanomalous regions are still assigned high anomaly scores, which has an adverse effect on the inspection accuracy. The US method yields reasonable results, but it is trained in a multiscale fashion, which requires an ensemble of 9 independent models, thus limiting its practical applications. Moreover, the detection effect varies greatly with the scale. Lv's method demonstrates robust performance on all types of textured surfaces.

Among all the compared methods, FMR-Net achieves the best comprehensive inspection accuracy on all types of defects and textures. As shown in Table \ref{table4}, compared to the second-best results, FMR-Net improves the AUC ROC values by margins of 0.1, 0.4, 0.2, and 2.6 on four types of textured surfaces and only slightly underperforms compared to RIAD on the leather dataset. Notably, unlike existing methods such as RIAD, our method demonstrates robust performance for all types of textured surfaces. The performance of the proposed method is essentially the same as that of state-of-the-art methods on simple textures, while for the most complex and irregular surface type, the performance of FMR-Net on the tile dataset is greatly improved. This finding can be attributed to the stable defect repair and background reconstruction of our FMR-Net.

Fig. \ref{fig11} presents the qualitative results of FMR-Net on the MVTec AD texture dataset, which illustrates its remarkable capability of simultaneous detection for various materials.

\subsubsection{Comparative results for the defect binarization detection task} In some industrial scenarios, it is often necessary to make binary judgments regarding the existence of defects on product surfaces. To verify the performance with respect to this requirement, we conducted comparative experiments on multiple surfaces in the Mvtech AD and DAGM datasets using several state-of-the-art models: AE\cite{r46}, AE-SSIM\cite{r46}, OC-GAN\cite{r50}, MS-FACE\cite{r9}, GPND\cite{r55}, and AFEAN\cite{r29}. Compared to the anomaly detection task addressed previously, this task requires the extra step of segmenting the anomaly map using the k-sigma threshold method\cite{r29}. The defects in the DAGM dataset are especially challenging to detect due to low contrast (such as in wallpaper and fabric samples) and large scale/orientation variations (such as in cement and MAGtile samples), which place great demands on inspection performance.

The qualitative comparative results are exhibited in Fig. \ref{fig15}. As shown, for this category of samples that are difficult to inspect, most existing methods, such as AE, AE-SSIM, OC-GAN and GPND, show inadequate performance. MS-FACE obtains good performance on specific textures, such as fabric and MAGtile. However, it is incapable of simultaneously detecting all various types of texture defects. AFEAN handles all types of defects and texture surfaces well, and our FMR-Net is also effective for defect detection in all types of textured surfaces.

The quantitative comparative results of three metrics, precision, recall and F1-measure, are reported in Table \ref{table5}. The proposed FMR-Net method is superior in terms of \emph{Precision} on all types of texture samples and ranks second only to AFEAN in \emph{Recall}'s performance. This result occurs because FMR-Net tends to yield more accurate detection results by means of the multimodal inspection method, avoiding introducing false detection noise. Thus, a higher $Precision$ but a reduced $Recall$ are produced. For comprehensive performance, from the comparison, it can be seen that FMR-Net improves the $F$1-measure value on most textures or is slightly behind the existing best method. However, for MAGtile samples with microdefects, there is a relatively large gap between FMR-Net and AFEAN due to the insensitive response of a certain mode to such defects, which will be further investigated in our future studies.

\subsection{Multilevel detection method towards practical application} The RSDDs data set\cite{r45} was collected from the surfaces of in-service rails, which suffer from diverse working conditions and nonuniform illumination conditions. More importantly, for slender workpieces such as rails, with the extreme imbalance between defect and background regions, unnecessary computational resources spent on normal areas will greatly limit efficiency. To address these challenges, we propose a multilevel detection method for RSDDs that enables fast inspection of slender rails for anomalous regions in a joint-domain fashion.

\begin{figure*}[t]
\centerline{\includegraphics[width=\textwidth]{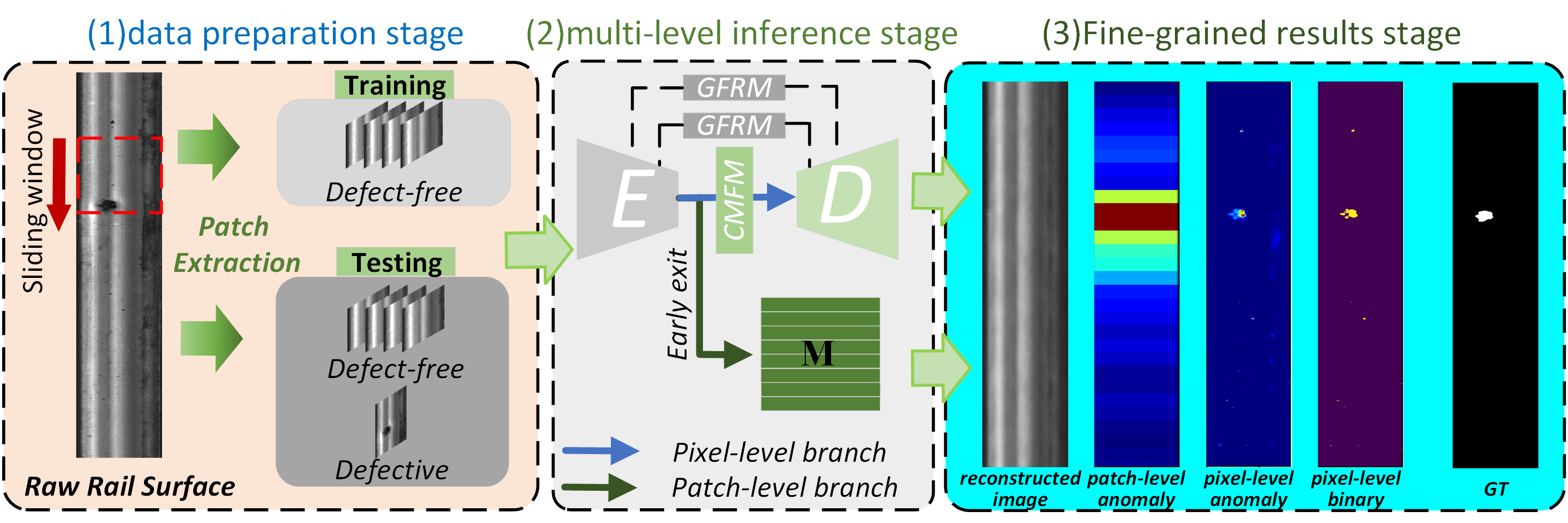}}
\caption{
The multilevel detection method on the RSDDs data set. The raw rail surface image is sliced into patches, and only some of the defect-free patches are utilized for model training, while the other defect-free patches and all of the defective patches are employed for model validation. During the multilevel inference phase, FMR-Net provides dual options for different fine-grained results.
}
\label{fig16}
\end{figure*}

\begin{figure}[t]
\centerline{\includegraphics[width=\columnwidth]{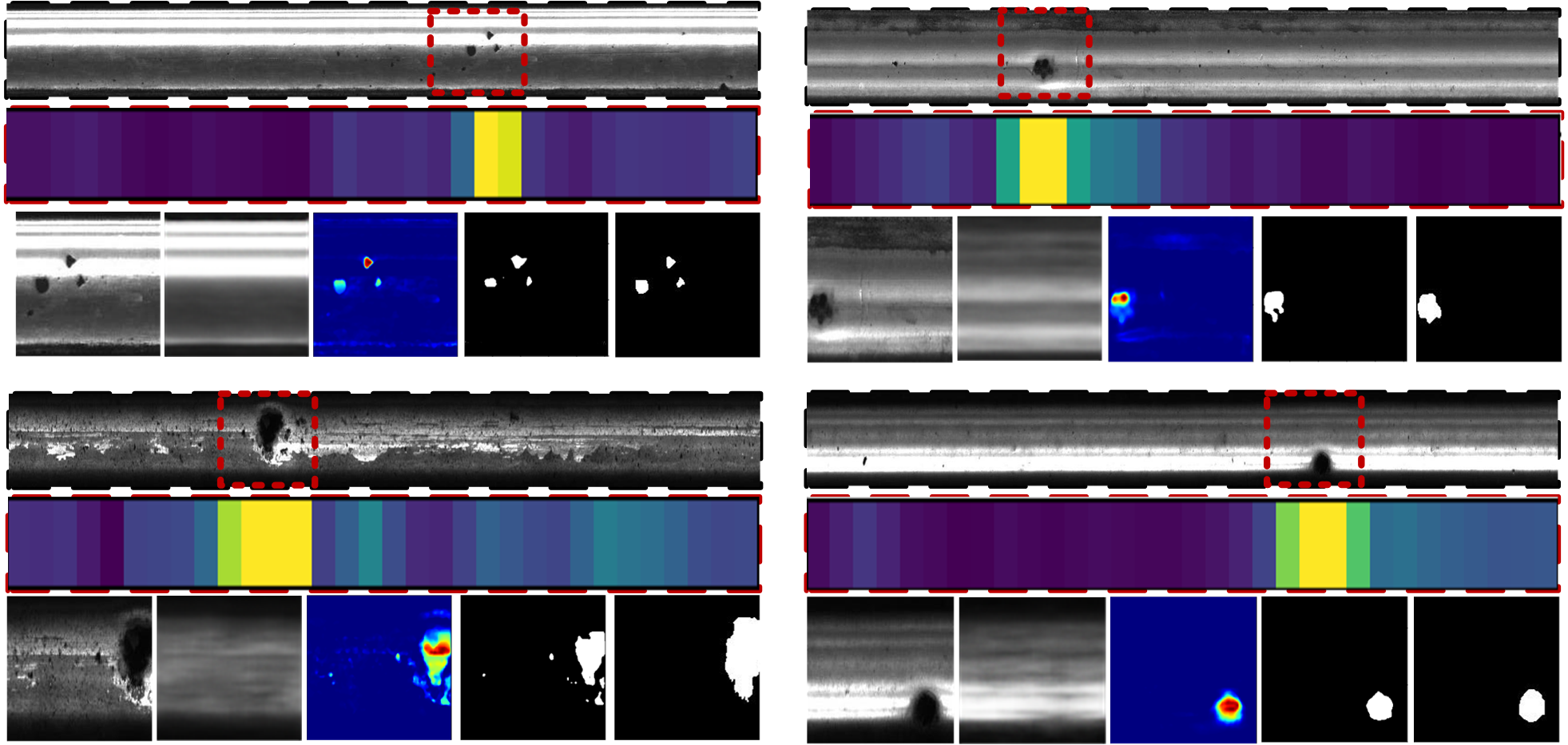}}
\caption{
Examples of the defect inspection performance of FMR-Net on the RSDDs dataset. First row: raw rail surfaces. Second row: patch-level inspection results. Third row: pixel-level inspection results. From left to right are the input defective images, the reconstructed images, the pixel-level anomaly results, the pixel-level binary results and the ground truth.}
\label{fig17}
\end{figure}

The overall schema is illustrated in Fig. \ref{fig16}. During the data preparation stage, the raw images are sliced into patches to constitute the training set and test set. As shown in Fig. \ref{fig16} (2), we propose multilevel inference via early exit from the encoder network. The proposed FMR-Net is additionally augmented with a side branch in the latent space. This architecture allows the inspection results for a large portion of defect-free samples to exit the model early via this branch when the fine-grained inspection requirements are already met. We exploit the observation that low-level fine-grained result may often be sufficient in some real industrial scenarios for the purpose of efficiency. In particular, we predefine two sets of fine-grained detection criteria, referred to as patch-level and pixel-level criteria, as shown in Fig. \ref{fig16} (3).

From the full inference branch, pixel-level inspection results can be obtained. In accordance with the methods described previously, we obtain the results of anomaly inspection or binarization inspection at the pixel level. From the early exit branch \cite{r21}, patch-level inspection results can be obtained. In detail, the encoder takes each patch as input and computes its latent representation. As mentioned in Section III-C, the features of typical normal patterns are recorded in the memory bank $M$ and can be employed as a reference for abnormality. The Euclidean distance to the nearest neighboring entry in the memory bank is defined as the anomaly score of a patch.

Examples of the inspection results are illustrated in Fig. \ref{fig17}. For patch-level anomaly localization, the proposed patch-level method requires approximately 1--2 ms, significantly improving the detection efficiency compared with the pixel-level method, which requires 7--8 ms; this efficiency is suitable for a case with a marked imbalance between defect and background regions.

For comparison with the competing methods AnoGAN\cite{r18}, OCGAN\cite{r50}, VAE\cite{r52}, MemAE\cite{r23}, GANomaly\cite{r30}, and DPAE\cite{r53}, the evaluation metrics are presented in Table \ref{table6}. These results show that our method achieves accurate defect detection at the pixel level. In terms of the AUC ROC and $precision$ evaluation indicator, our method is superior to the other methods, while the $recall$ and $F1$-measures are slightly lower than those of GANomaly.

\begin{table}
\caption{Performance comparison of the proposed method and other compared schemes on the RSDDs data set}
\label{table}
\setlength{\tabcolsep}{3pt}
\begin{tabular}{p{\columnwidth}}
$\includegraphics[width=\columnwidth]{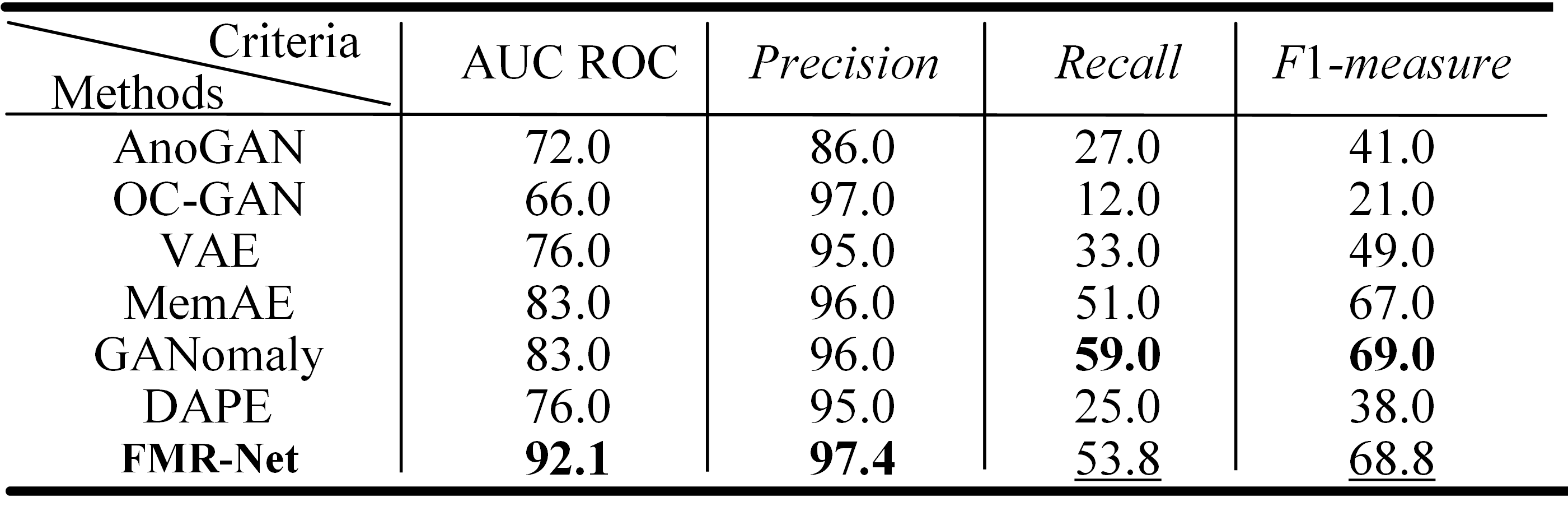}$
\end{tabular}
\label{table6}
\end{table}
In summary, all of the above experimental results demonstrate that the proposed FMR-Net method achieves comprehensively superior performance in online industrial inspection, making it suitable for deployment in edge-intelligence-enabled intelligent manufacturing scenarios.

\section{Industrial Application in the Edge Intelligence Paradigm}

\begin{figure}[t]
\centerline{\includegraphics[width=\columnwidth]{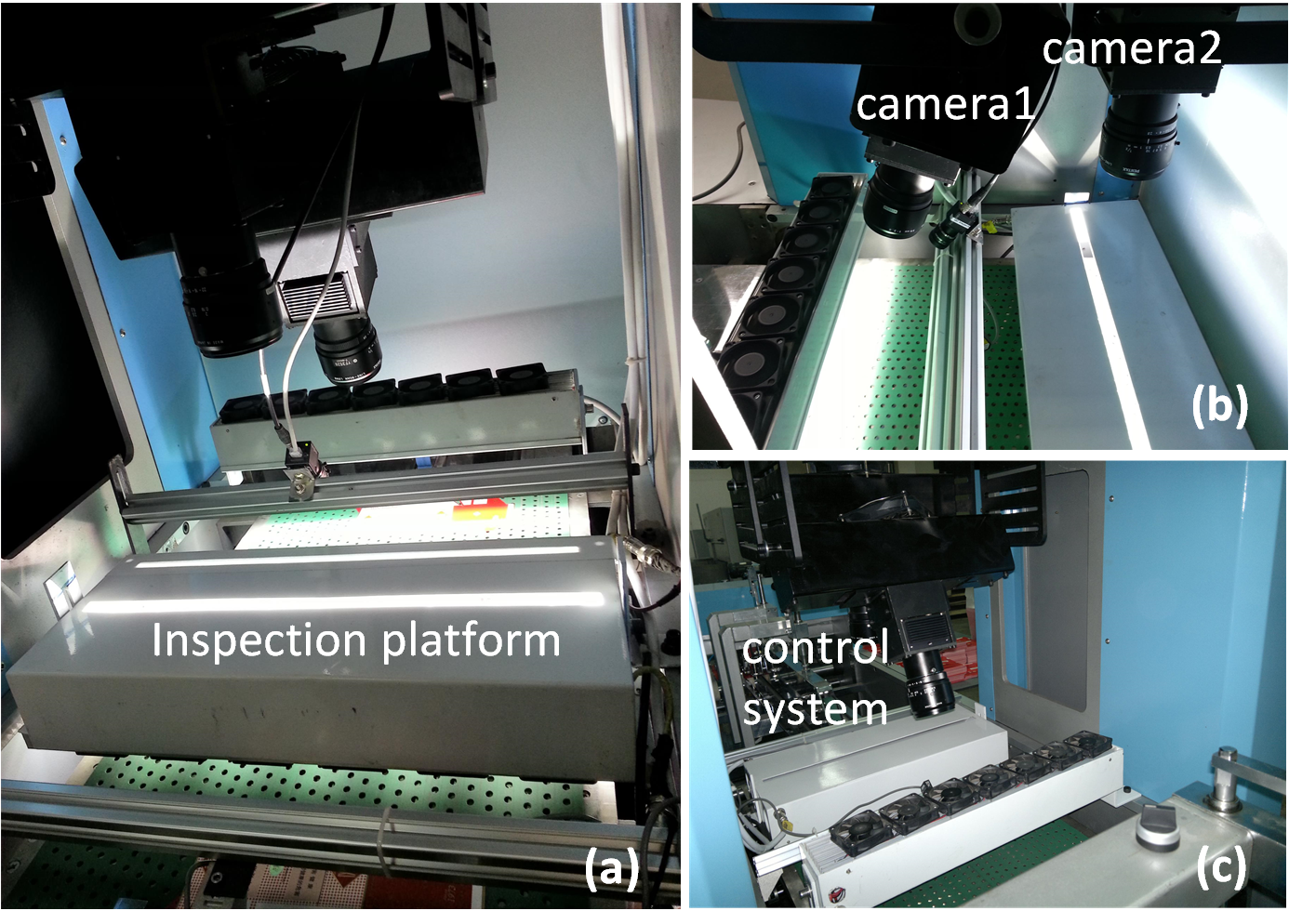}}
\caption{AOI equipment for PPS defect inspection. (a) Inspection platform for PPS defects.
(b) Multiview imaging system. (c) Control system for the whole device.}
\label{fig18}
\end{figure}

For the further practical evaluation of FMR-Net, it was implemented in our AOI equipment and deployed for the online inspection of printed product surface (PPS) defects, as shown in Fig. \ref{fig18}. Our AOI equipment mainly consists of a multiview camera, an inspection platform and a control system and can provide premium imaging performance for textured surfaces of wide breadth. A raw panel has a high resolution of 2560×3840 pixels, and many microdefects may become attached along the production line. These defects come in an endless variety of types, and it would be unrealistic to collect sufficient samples of all of them. Thus, there are strict limitations on the capabilities of supervised methods.

With the revolution of intelligent manufacturing, many IIoT devices equipped with artificial intelligence (AI) algorithms have emerged. Conventional cloud-based approaches usually involve offloading workloads to backbone data center servers; however, they incur significant latency,
communication overhead, and severe packet loss caused by unstable communication bandwidth, which may influence the production speed. To overcome these issues, edge computing\cite{r20} is emerging as a new paradigm of interest in the field of intelligent manufacturing inspection\cite{r32,r33,r34}. Accordingly, we implemented the proposed inspection method in practice with the aid of edge computing.

\begin{figure}[t]
\centerline{\includegraphics[width=\columnwidth]{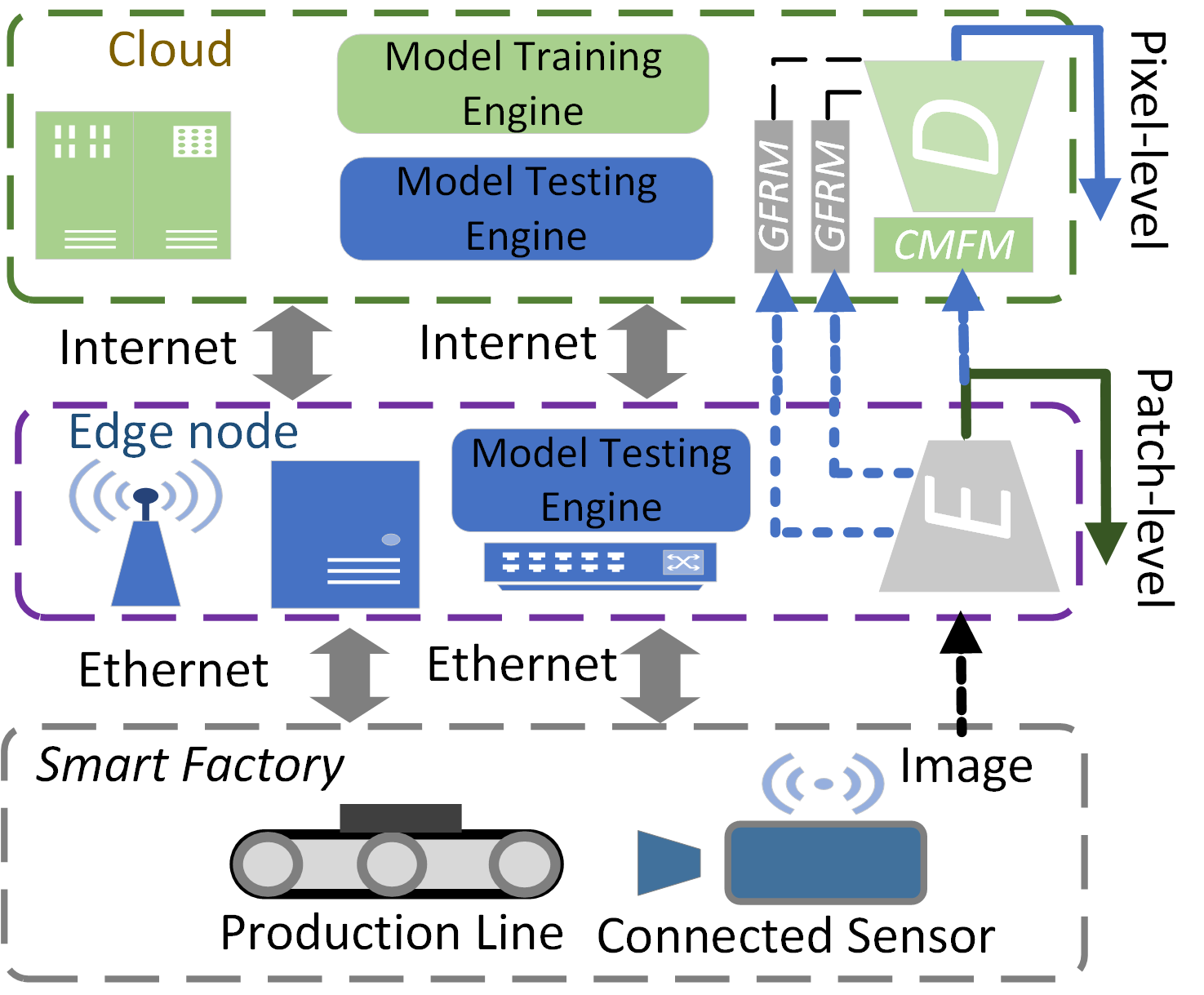}}
\caption{Overview of the cloud--edge-collaboration-enabled online defect inspection system.}
\label{fig19}
\end{figure}

The overall system framework in the cloud--edge collaborative computational paradigm proposed here is depicted in Fig. \ref{fig19}. Image sensors are mounted along the production line to monitor the manufacturing status, and the cameras are connected to the network using a highly reliable transmission channel, such as Ethernet or 5G, in order to upload the image data to local edge nodes. Our online inspection phase, in which quick response is needed, is partially implemented at the edge nodes. Considering the multilevel inspection method, we adopt the model partitioning technique\cite{r22} as a way to allocate lower layers (the encoder) to the edge nodes while ensuring that early patch-level results can be obtained without waiting for the full inference results from the cloud server. If the patch-level results already meet the requirements, we can stop uploading the intermediate feature maps and adopt the fast inference results as the final results. Otherwise, the intermediate feature maps should be uploaded to the central cloud server for the further computation of pixel-level results. In this way, the fast inference results can directly inform the manufacturing process on site. With respect to the model training, the offline computational tasks, which is computationally heavy and does not require a quick response. This processing is naturally deployed on the cloud server, which has strong computational capabilities.

The detailed setup of the hardware is presented as follows: The edge nodes are equipped with the NVIDIA Jetson AGX Xavier computing platform under the Ubuntu 16.04 operating system, and the algorithms are implemented in the Python 3.8 environment, mainly based on the open-source TensorFlow package. The backbone cloud center is composed of a server with 4 Nvidia 2080 Ti GPU cards.

\begin{figure}[t]
\centerline{\includegraphics[width=\columnwidth]{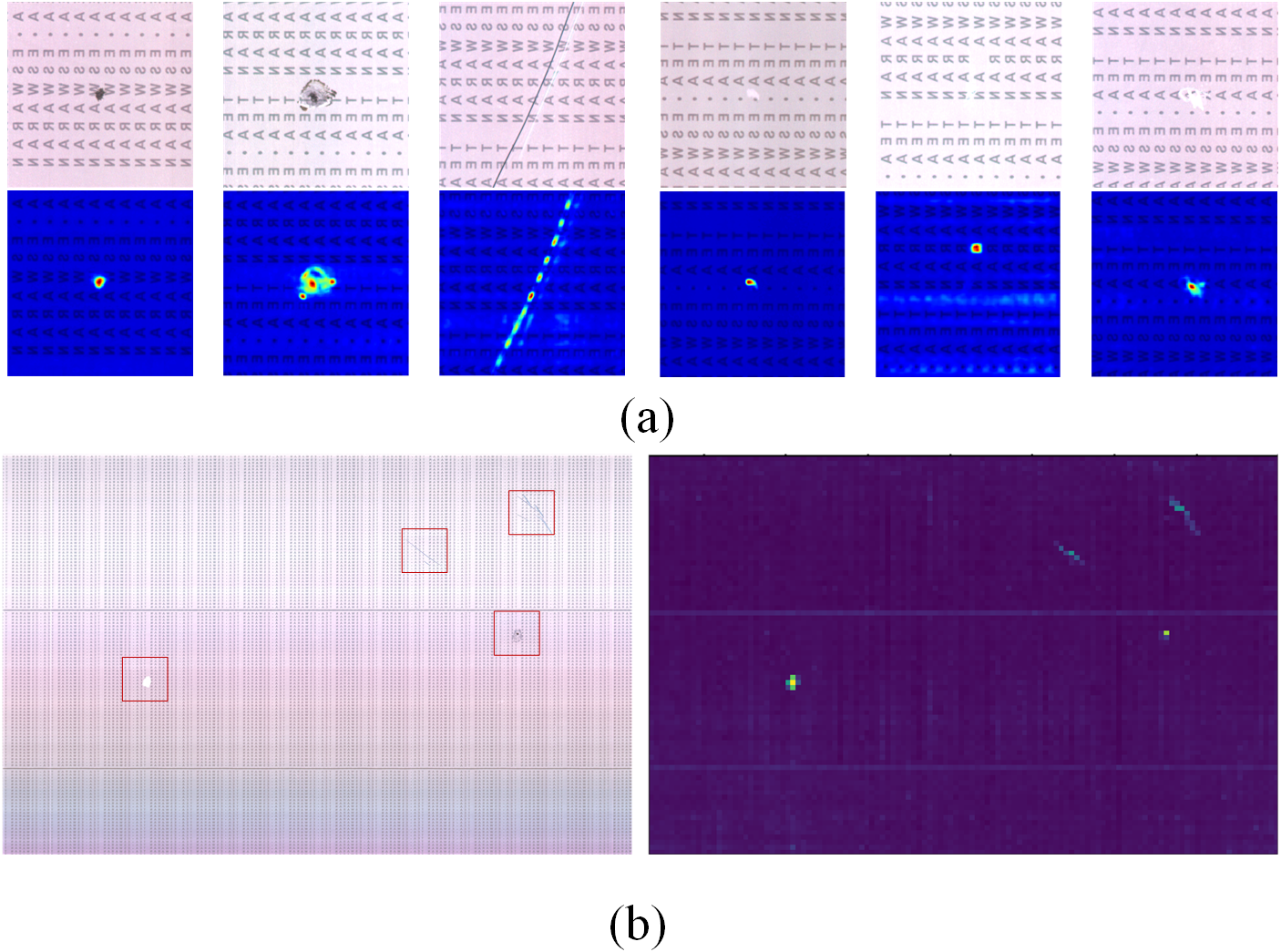}}
\caption{Defect inspection results obtained for PPS images using our proposed FMR-Net implementation. (a) Pixel-level inspection results: original samples (first row) and samples with inspection results overlaid (second row). From left to right, the samples show a spot defect, a stain defect, a scratch defect, a fading defect, a missing print defect and a dye defect. (b) Raw wide-breadth PPS defect image and its patch-level inspection results.}
\label{fig20}
\end{figure}

Some examples of PPS defects and corresponding pixel-level inspection results are shown in Fig. \ref{fig20} (a). The original images and the images with the detection results overlaid are shown in the first and second rows, respectively. The inspection results show that FMR-Net can both simultaneously and accurately inspect products for all types of PPS defects, regardless of their color, shape or scale. An example of a raw high-resolution defective PPS image and its patch-level inspection results is displayed in Fig. \ref{fig20} (b), from which we can conclude that the coarse localization of defects is achieved, showing great potential for use in high-speed industrial inspection.

\section{Conclusion and Discussion}

In this paper, we propose a novel method, FMR-Net, for the visual inspection of textured surfaces for defects in an unsupervised fashion. This method requires only defect-free and synthetic defective samples and does not require any real, manually labeled defect samples. The proposed FMR-Net relies on feature memory rearrangement to restore the features of defective image regions. First, an encoding module is employed to obtain multiscale features for textured surfaces. Then, the proposed CMFM is used to improve the discriminability of the latent features and substitute normal features for anomalous features. Next, the proposed GFRM is utilized for residual defect suppression on the skip connection pathways. Thus, the decoding module can exploit the restored features to reconstruct the texture background. Furthermore, a two-phase training strategy and a multimodal inspection method are proposed to improve the restoration capability and detection accuracy of the model. Extensive experimental results on several textured surface data sets demonstrate that FMR-Net exhibits state-of-the-art overall inspection performance and confirm the contributions of the various modules and operational processes.

However, due to the high resolution of the original images, multiple patches must be extracted to train FMR-Net. Unfortunately, in contrast to texture images, patches cut from object images are highly relevant to the location distribution and contain complex semantic information, which is difficult to learn due to the limited receptive field. In the future, we will investigate advanced patch extraction strategies to further improve this approach.

Finally, we have proposed a multilevel detection method for scenarios of defect/background imbalance and have deployed it in a practical online inspection application within the edge--cloud collaborative paradigm. Currently, with the development of IIoT technology, edge computing is showing great potential for relieving the computational and communication burdens arising in various industrial scenarios. In the future, we will also direct more attention toward edge-computing-enabled inspection applications.

\bibliographystyle{ieeetr}
\bibliography{reference}

\begin{thebibliography}{10}

\bibitem{r1}
B.~Mallik-Goswami and A.~K. Datta, ``Detecting defects in fabric with
  laser-based morphological image processing,'' {\em Textile Research Journal},
  vol.~70, no.~9, pp.~758--762, 2000.

\bibitem{r2}
H.~Y. Wang, J.~Zhang, Y.~Tian, H.~Y. Chen, H.~X. Sun, and K.~Liu, ``A simple
  guidance template-based defect detection method for strip steel surfaces,''
  {\em IEEE Transactions on Industrial Informatics}, pp.~1--1, 2018.

\bibitem{r3}
H.~Yang, K.~Song, S.~Mei, and Z.~Yin, ``An accurate mura defect vision
  inspection method using outlier-prejudging-based image background
  construction and region-gradient-based level set,'' {\em IEEE Transactions on
  Automation Science and Engineering}, pp.~1--18, 2018.

\bibitem{r4}
Bae-Keun, Kwon, Jong-Seob, Won, Dong-Joong, and Kang, ``Fast defect detection
  for various types of surfaces using random forest with vov features,'' {\em
  International Journal of Precision Engineering \&amp; Manufacturing}, 2015.

\bibitem{r5}
B.~Zhang, H.~Yang, and Z.~Yin, ``A region-based normalized cross correlation
  algorithm for the vision-based positioning of elongated ic chips,'' {\em IEEE
  Transactions on Semiconductor Manufacturing}, vol.~28, no.~3, pp.~345--352,
  2015.

\bibitem{r6}
H.~Yang, S.~Mei, K.~Song, B.~Tao, and Z.~Yin, ``Transfer-learning-based online
  mura defect classification,'' {\em IEEE Transactions on Semiconductor
  Manufacturing}, vol.~31, no.~1, pp.~116--123, 2018.

\bibitem{r27}
S.~Niu, B.~Li, X.~Wang, and H.~Lin, ``Defect image sample generation with gan
  for improving defect recognition,'' {\em IEEE Transactions on Automation
  Science and Engineering}, vol.~17, no.~3, pp.~1611--1622, 2020.

\bibitem{r7}
Y.~He, K.~Song, Q.~Meng, and Y.~Yan, ``An end-to-end steel surface defect
  detection approach via fusing multiple hierarchical features,'' {\em IEEE
  Transactions on Instrumentation and Measurement}, vol.~PP, no.~99, pp.~1--1,
  2019.

\bibitem{r8}
G.~Song, K.~Song, and Y.~Yan, ``Edrnet: Encoder–decoder residual network for
  salient object detection of strip steel surface defects,'' {\em IEEE
  Transactions on Instrumentation and Measurement}, vol.~PP, no.~99, pp.~1--1,
  2020.

\bibitem{r54}
X.~Dong, C.~J. Taylor, and T.~F. Cootes, ``Defect classification and detection
  using a multitask deep one-class cnn,'' {\em IEEE Transactions on Automation
  Science and Engineering}, pp.~1--12, 2021.

\bibitem{r9}
H.~Yang, Y.~Chen, K.~Song, and Z.~Yin, ``Multiscale feature-clustering-based
  fully convolutional autoencoder for fast accurate visual inspection of
  texture surface defects,'' {\em IEEE Transactions on Automation Science and
  Engineering}, pp.~1--18, 2019.

\bibitem{r10}
T.~Vujasinovic, J.~Pribic, K.~Kanjer, N.~T. Milosevic, Z.~Tomasevic,
  Z.~Milovanovic, D.~Nikolic-Vukosavljevic, and M.~Radulovic, ``Gray-level
  co-occurrence matrix texture analysis of breast tumor images in prognosis of
  distant metastasis risk,'' {\em Microscopy \& Microanalysis}, vol.~21,
  no.~03, pp.~646--654, 2015.

\bibitem{r11}
N.~Sharma, A.~K. Ray, S.~Sharma, K.~K. Shukla, and L.~M. Aggarwal,
  ``Segmentation and classification of medical images using texture-primitive
  features: Application of bam-type artificial neural network,'' {\em Journal
  of Medical Physics}, vol.~33, no.~3, pp.~119--126, 2008.

\bibitem{r12}
D.~M. Tsai and T.~Y. Huang, ``Automated surface inspection for statistical
  textures,'' {\em IMAGE AND VISION COMPUTING}, 2003.

\bibitem{r13}
X.~Xie and M.~Mirmehdi, ``Texems: Texture exemplars for defect detection on
  random textured surfaces,'' {\em IEEE Transactions on Pattern Analysis \&
  Machine Intelligence}, vol.~29, no.~8, pp.~1454--1464, 2007.

\bibitem{r14}
D.~Aiger and H.~Talbot, ``The phase only transform for unsupervised surface
  defect detection,'' in {\em Computer Vision \& Pattern Recognition}, 2010.

\bibitem{r15}
D.~Tabernik, S.~Ela, J.~Skvar, and D.~Skoaj, ``Segmentation-based deep-learning
  approach for surface-defect detection,'' {\em Journal of Intelligent
  Manufacturing}, vol.~31, 2020.

\bibitem{r16}
H.~Dong, K.~Song, Y.~He, J.~Xu, Y.~Yan, and Q.~Meng, ``Pga-net: Pyramid feature
  fusion and global context attention network for automated surface defect
  detection,'' {\em IEEE Transactions on Industrial Informatics}, vol.~PP,
  no.~99, pp.~1--1, 2020.

\bibitem{r17}
S.~Mei, H.~Yang, and Z.~Yin, ``An unsupervised-learning-based approach for
  automated defect inspection on textured surfaces,'' {\em IEEE Transactions on
  Instrumentation and Measurement}, pp.~1266--1277, 2018.

\bibitem{r18}
T.~Schlegl, P.~Seebck, S.~M. Waldstein, U.~Schmidt-Erfurth, and G.~Langs,
  ``Unsupervised anomaly detection with generative adversarial networks to
  guide marker discovery,'' {\em Springer, Cham}, 2017.

\bibitem{r19}
Z.~Zhou, X.~Chen, E.~Li, L.~Zeng, K.~Luo, and J.~Zhang, ``Edge intelligence:
  Paving the last mile of artificial intelligence with edge computing,'' {\em
  Proceedings of the IEEE}, 2019.

\bibitem{r20}
W.~Shi, C.~Jie, Z.~Quan, Y.~Li, and L.~Xu, ``Edge computing: Vision and
  challenges,'' {\em Internet of Things Journal, IEEE}, vol.~3, no.~5,
  pp.~637--646, 2016.

\bibitem{r21}
S.~Teerapittayanon, B.~McDanel, and H.~Kung, ``Branchynet: Fast inference via
  early exiting from deep neural networks,'' {\em 2016 23rd International
  Conference on Pattern Recognition (ICPR)}, pp.~2464--2469, 2016.

\bibitem{r22}
Y.~Kang, J.~Hauswald, G.~Cao, A.~Rovinski, T.~Mudge, J.~Mars, L.~Tang, and
  C.~Lab, ``Neurosurgeon: Collaborative intelligence between the cloud and
  mobile edge,'' {\em Acm Sigplan Notices}, vol.~52, no.~1, pp.~615--629, 2017.

\bibitem{r23}
D.~Gong, L.~Liu, V.~Le, B.~Saha, M.~R. Mansour, S.~Venkatesh, and A.~Hengel,
  ``Memorizing normality to detect anomaly: Memory-augmented deep autoencoder
  for unsupervised anomaly detection,'' in {\em 2019 IEEE/CVF International
  Conference on Computer Vision (ICCV)}, 2020.

\bibitem{r24}
A.~Vz, A.~Mk, and A.~Ds, ``Reconstruction by inpainting for visual anomaly
  detection,'' {\em Pattern Recognition}, 2020.

\bibitem{r25}
G.~Liu, R.~Taori, T.~Wang, Z.~Yu, S.~Liu, F.~A. Reda, K.~Sapra, A.~Tao, and
  B.~Catanzaro, ``Transposer: Universal texture synthesis using feature maps as
  transposed convolution filter,'' {\em CoRR}, vol.~abs/2007.07243, 2020.

\bibitem{r26}
T.~Karras, S.~Laine, and T.~Aila, ``A style-based generator architecture for
  generative adversarial networks,'' in {\em 2019 IEEE/CVF Conference on
  Computer Vision and Pattern Recognition (CVPR)}, 2019.

\bibitem{r28}
C.~Lv, F.~Shen, Z.~Zhang, D.~Xu, and Y.~He, ``A novel pixel-wise defect
  inspection method based on stable background reconstruction,'' {\em IEEE
  Transactions on Instrumentation and Measurement}, vol.~PP, no.~99, pp.~1--1,
  2020.

\bibitem{r29}
H.~Yang, Q.~Zhou, K.~Song, and Z.~Yin, ``An anomaly-feature-editing-based
  adversarial network for texture defect visual inspection,'' {\em IEEE
  Transactions on Industrial Informatics}, vol.~PP, no.~99, pp.~1--1, 2020.

\bibitem{r30}
S.~Akcay, A.~Atapour-Abarghouei, and T.~P. Breckon, ``Ganomaly: Semi-supervised
  anomaly detection via adversarial training,'' in {\em Computer Vision -- ACCV
  2018} (C.~V. Jawahar, H.~Li, G.~Mori, and K.~Schindler, eds.), (Cham),
  pp.~622--637, Springer International Publishing, 2019.

\bibitem{r31}
S.~Akay, A.~Atapour-Abarghouei, and T.~P. Breckon, ``Skip-ganomaly: Skip
  connected and adversarially trained encoder-decoder anomaly detection,'' in
  {\em 2019 International Joint Conference on Neural Networks (IJCNN)}, 2019.

\bibitem{r32}
L.~Li, K.~Ota, and M.~Dong, ``Deep learning for smart industry: Efficient
  manufacture inspection system with fog computing,'' {\em IEEE Transactions on
  Industrial Informatics}, vol.~14, no.~10, pp.~4665--4673, 2018.

\bibitem{r33}
L.~Zeng, E.~Li, Z.~Zhou, and X.~Chen, ``Boomerang: On-demand cooperative deep
  neural network inference for edge intelligence on the industrial internet of
  things,'' {\em IEEE Network}, vol.~33, no.~5, pp.~96--103, 2019.

\bibitem{r34}
Y.~Wang, L.~Gao, P.~Zheng, H.~Yang, and J.~Zou, ``A smart surface inspection
  system using faster r-cnn in cloud-edge computing environment,'' {\em
  Advanced Engineering Informatics}, vol.~43, 2020.

\bibitem{r35}
Y.~Wang, K.~Hong, J.~Zou, T.~Peng, and H.~Yang, ``A cnn-based visual sorting
  system with cloud-edge computing for flexible manufacturing systems,'' {\em
  IEEE Transactions on Industrial Informatics}, vol.~PP, no.~99, pp.~1--1,
  2019.

\bibitem{r36}
J.~Yi and S.~Yoon, ``Patch svdd: Patch-level svdd for anomaly detection and
  segmentation,'' in {\em Computer Vision -- ACCV 2020} (H.~Ishikawa, C.-L.
  Liu, T.~Pajdla, and J.~Shi, eds.), (Cham), pp.~375--390, Springer
  International Publishing, 2021.

\bibitem{r37}
P.~Napoletano, F.~Piccoli, and R.~Schettini, ``Anomaly detection in nanofibrous
  materials by cnn-based self-similarity,'' {\em Sensors}, vol.~18, no.~1,
  2018.

\bibitem{r38}
J.~Yu, Z.~Lin, J.~Yang, X.~Shen, X.~Lu, and T.~S. Huang, ``Generative image
  inpainting with contextual attention,'' {\em IEEE}, 2018.

\bibitem{r39}
S.~C. Zhu, C.~E. Guo, Y.~Wang, and Z.~Xu, ``What are textons?,'' {\em
  International Journal of Computer Vision}, vol.~62, no.~1-2, pp.~121--143,
  2005.

\bibitem{r40}
W.~Xue, L.~Zhang, X.~Mou, and A.~C. Bovik, ``Gradient magnitude similarity
  deviation: A highly efficient perceptual image quality index,'' {\em IEEE
  Transactions on Image Processing}, vol.~23, no.~2, pp.~684--695, 2014.

\bibitem{r41}
Hang, Zhao, Orazio, Gallo, Iuri, Frosio, Jan, and Kautz, ``Loss functions for
  image restoration with neural networks,'' {\em IEEE Transactions on
  Computational Imaging}, 2017.

\bibitem{r43}
P.~Bergmann, M.~Fauser, D.~Sattlegger, and C.~Steger, ``Mvtec ad — a
  comprehensive real-world dataset for unsupervised anomaly detection,'' in
  {\em 2019 IEEE/CVF Conference on Computer Vision and Pattern Recognition
  (CVPR)}, 2020.

\bibitem{r44}
M.~Wieler and T.~Hahn, ``Weakly supervised learning for industrial optical
  inspection abstract.''
  \url{https://hci.iwr.uni-heidelberg.de/content/weakly-supervised-learning-industrial-optical-inspection}.

\bibitem{r45}
H.~Yu, Q.~Li, Y.~Tan, J.~Gan, J.~Wang, Y.~A. Geng, and L.~Jia, ``A
  coarse-to-fine model for rail surface defect detection,'' {\em IEEE
  Transactions on Instrumentation and Measurement}, vol.~PP, no.~3, pp.~1--11,
  2018.

\bibitem{r46}
P.~Bergmann, S.~Lwe, M.~Fauser, D.~Sattlegger, and C.~Steger, ``Improving
  unsupervised defect segmentation by applying structural similarity to
  autoencoders,'' in {\em 14th International Conference on Computer Vision
  Theory and Applications}, 2019.

\bibitem{r47}
T.~Schlegl, P.~Seebck, S.~M. Waldstein, G.~Langs, and U.~Schmidt-Erfurth,
  ``f-anogan: Fast unsupervised anomaly detection with generative adversarial
  networks,'' {\em Medical Image Analysis}, vol.~54, 2019.

\bibitem{r48}
P.~Bergmann, M.~Fauser, D.~Sattlegger, and C.~Steger, ``Uninformed students:
  Student-teacher anomaly detection with discriminative latent embeddings,''
  {\em IEEE}, 2020.

\bibitem{r50}
P.~Perera, R.~Nallapati, and X.~Bing, ``Ocgan: One-class novelty detection
  using gans with constrained latent representations,'' {\em IEEE}, 2019.

\bibitem{r55}
S.~Pidhorskyi, R.~Almohsen, D.~A. Adjeroh, and G.~Doretto, ``Generative
  probabilistic novelty detection with adversarial autoencoders,'' in {\em
  Proceedings of the 32nd International Conference on Neural Information
  Processing Systems}, NIPS'18, (Red Hook, NY, USA), p.~6823–6834, Curran
  Associates Inc., 2018.

\bibitem{r52}
D.~P. Kingma and M.~Welling, ``Auto-encoding variational bayes,'' {\em arXiv
  preprint arXiv:1312.6114}, 2013.

\bibitem{r53}
J.~Liu, K.~Song, M.~Feng, Y.~Yan, and L.~Zhu, ``Semi-supervised anomaly
  detection with dual prototypes autoencoder for industrial surface
  inspection,'' {\em Optics and Lasers in Engineering}, vol.~136, p.~106324,
  2021.

\end{thebibliography}

\begin{IEEEbiography}[{\includegraphics[width=1in,height=1.25in,clip,keepaspectratio]{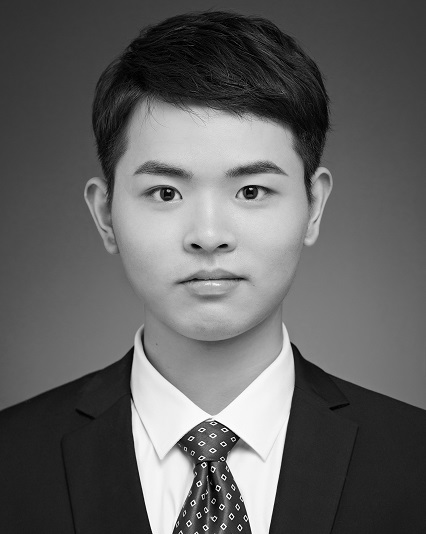}}]{Haiming Yao} received a B.S. degree from the School of Mechanical Science and Engineering, Huazhong University of Science and Technology, Wuhan, China, in 2022.
He is pursuing a Ph.D. degree with
the Department of Precision Instrument, Tsinghua
University.

His research interests include deep learning, edge intelligence and machine vision.
\end{IEEEbiography}

\begin{IEEEbiography}[{\includegraphics[width=1in,height=1.25in,clip,keepaspectratio]{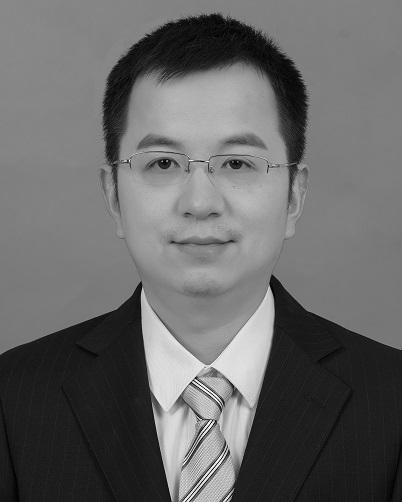}}]{Wenyong Yu} received an M.S. degree and a Ph.D. degree from Huazhong University of Science and Technology, Wuhan, China, in 1999 and 2004, respectively. He is currently an Associate Professor with the School of Mechanical Science and Engineering, Huazhong University of Science and Technology. 

His research interests include machine vision, intelligent control, and image processing.
\end{IEEEbiography}

\begin{IEEEbiography}[{\includegraphics[width=1in,height=1.25in,clip,keepaspectratio]{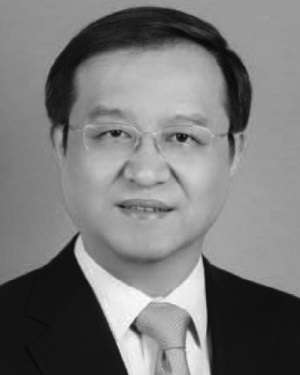}}]{Xue Wang}received an M.S. degree in measurement and instrumentation from Harbin Institute of Technology,
Harbin, China, in 1991 and a Ph.D. degree in mechanical engineering from Huazhong University of Science and Technology, Wuhan, China, in 1994.

He was a Postdoctoral Fellow in electrical power systems with Huazhong University of Science and Technology from 1994 to 1996. He then joined
the Department of Precision Instrument, Tsinghua University, Beijing, China, where he is currently a Professor. From May 2001 to July 2002, he was a Visiting Professor with the Department of Mechanical Engineering, University of Wisconsin--Madison. His research interests include topics in wireless sensor networks, cyber-physical systems, intelligent biosignal processing, medical image processing, and smart energy utilization.

Prof. Wang is a senior member of the IEEE Instrumentation and Measurement Society, Computer Society, Computational Intelligence Society, and Communications Society.
\end{IEEEbiography}

\end{document}